\documentclass[final]{cvpr}
\usepackage{graphicx}
\usepackage{comment}
\usepackage{amsmath,amssymb,amsfonts} %

\usepackage{mathtools}
\usepackage{url}
\usepackage{bm}
\usepackage{wrapfig}
\usepackage{multirow}
\usepackage[dvipsnames]{xcolor}
\usepackage{newtxtext}
\usepackage{subfig}
\usepackage{enumitem}
\usepackage[normalem]{ulem}
\useunder{\uline}{\ul}{}
\usepackage{cite}

\usepackage{adjustbox}

\usepackage[pagebackref=true,breaklinks=true,colorlinks,bookmarks=false]{hyperref}
\usepackage{cleveref}

\makeatletter
\@namedef{ver@everyshi.sty}{}
\makeatother
\usepackage{tikz}
\usetikzlibrary{arrows}

\usepackage[font=small]{caption}

\renewcommand\paragraph[1]{\vspace{.6ex}\noindent\textbf{#1}}

\def\cvprPaperID{3192} %
\def\confYear{CVPR 2021}

\newcommand{\C}{\mathbf{C}}
\newcommand{\X}{\mathbf{X}}
\newcommand{\x}{\mathbf{x}}
\newcommand{\w}{\mathbf{w}}
\newcommand{\K}{\mathbf{K}}
\newcommand{\R}{\mathbf{R}}
\newcommand{\Rt}{\mathbf{R_t}}
\renewcommand{\P}{\mathbf{P}}
\renewcommand{\t}{\mathbf{t}}
\newcommand{\bbR}{\mathbb{R}}

\newcommand{\object}{\mathbb{O}}
\newcommand{\M}{\mathbf{M}}

\newcommand{\feature}{\mathcal{F}}

\newcommand{\V}{\mathbf{V}}
\renewcommand{\P}{\mathbf{P}}
\newcommand{\D}{\mathbf{D}}
\newcommand{\depthSet}{\mathcal{D}}

\newcommand{\modelname}{NeRD}

\begin{document}

\setlength{\belowcaptionskip}{0pt}
\setlength{\abovecaptionskip}{1.2ex}
\addtolength{\floatsep}{-1ex}
\addtolength{\textfloatsep}{-2ex}
\addtolength{\dbltextfloatsep}{-1.5ex}
\setlength{\abovedisplayskip}{6pt}
\setlength{\belowdisplayskip}{6pt}
\newcommand{\simplify}[1]{}

\title{NeRD: Neural 3D Reflection Symmetry Detector}

\author{%
  Yichao Zhou \\
  Univ. of California, Berkeley \\
  \texttt{zyc@berkeley.edu} \\
  \and
  Shichen Liu \\
  Univ. of Southern California \\
  \texttt{liushich@usc.edu} \\
  \and
  Yi Ma \\
  Univ. of California, Berkeley \\
  \texttt{yima@eecs.berkeley.edu} \\
}

\maketitle

\begin{abstract}
Recent advances have shown that symmetry, a structural prior that most objects exhibit, can support a variety of single-view 3D understanding tasks.
However, detecting 3D symmetry from an image remains a challenging task.
Previous works either assume that the symmetry is given or detect the symmetry with a heuristic-based method.
In this paper, we present \modelname{}, a \textbf{Ne}ural 3D \textbf{R}eflection Symmetry \textbf{D}etector, which combines the strength of learning-based recognition and geometry-based reconstruction to accurately recover the normal direction of objects' mirror planes. 
Specifically, we first enumerate the symmetry planes with a coarse-to-fine strategy and then find the best ones by building 3D cost volumes to examine the intra-image pixel correspondence from the symmetry.
Our experiments show that the symmetry planes detected with our method are significantly more accurate than the planes from direct CNN regression on both synthetic and real-world datasets.
We also demonstrate that the detected symmetry can be used to improve the performance of downstream tasks such as pose estimation and depth map regression.
The code of this paper has been made public at \url{https://github.com/zhou13/nerd}.
\end{abstract}

\section{Introduction}
Recovering the 3D orientation of objects in an image is a fundamental problem in 3D vision, which plays important roles in tasks such as robotics, autonomous driving, virtual reality (VR), augmented reality (AR), and 3D scene understanding. 
Traditionally, such a problem is hard to solve.  
Researchers can to RGB-D input captured with time-of-flight cameras or structured light \cite{choi20123d,song2016deep,qi2018frustum}.
Unfortunately, depth cameras often have limited range and can be interfered with by other light sources, and the requirement of owning a depth camera is inconvenient for average users, which severely restricts its applications.

Recent advances in convolutional neural networks in object detection and instance segmentation have shown good potential in inferring object-level information from RGB images by leveraging supervised learning.
Nowadays, single-view neural network-based methods are able to predict the object pose under different settings.
Some work explores the \emph{instance-level 3D pose estimation} problem \cite{tekin2018real,rad2017bb8,liu2019soft} in which the CAD models of the objects are known beforehand.
However, these settings are rather limited because in practice we do not have CAD models for many objects.
Therefore, other work tries to tackle the \emph{category-level 3D pose estimation} problem \cite{xiang2015data,chen2016monocular,mousavian20173d} without relying on the exact CAD models of objects.
Unlike the cases where either depth information or CAD models are available, previous single-view category-level 3D pose estimation methods can hardly exploit the geometric constraints between the input RGB image and the 3D shape and predict the pose solely by interpolating the training data.
Hence, such formulation is ill-posed, which leads to inaccurate pose recovery \cite{Equivalent2ImageClassification}.

 \begin{figure}[t]
  \centering
  \includegraphics[width=0.99\linewidth]{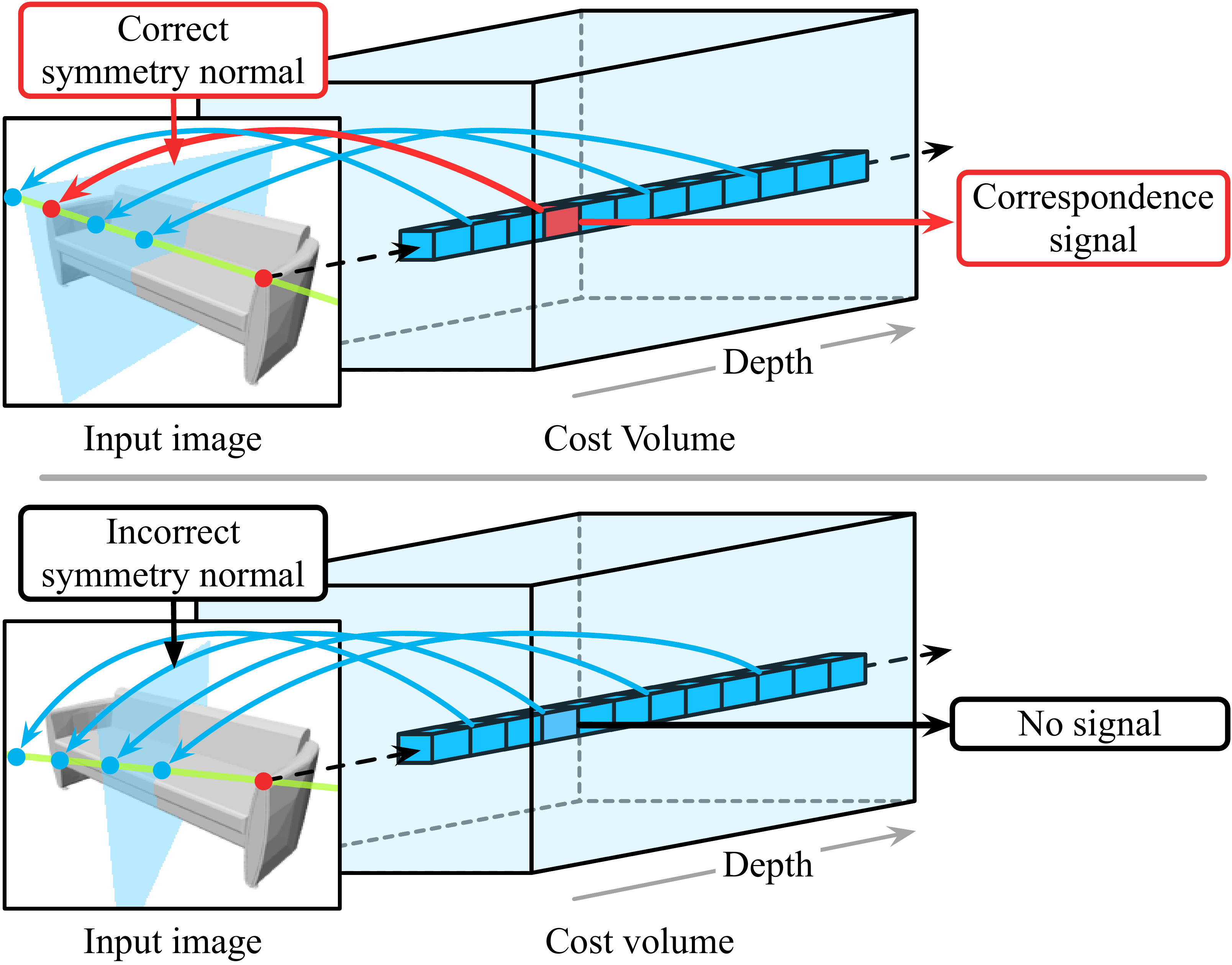}
  \caption{
  Illustration of the symmetry detection process in \modelname{}.  For each pixel, we enumerate its depth and warp features along the line according to the symmetry plane hypothesis.  If the hypothesis is correct, there should be matched features for most of the pixels.
  }
  \label{fig:teaser}
\end{figure}

To address this difficulty, we identify a structure that commonly exists in man-made objects, the \emph{reflection symmetry}, as a geometric connection between the object poses and the images.
We observe that the canonical space of objects often is determined by aligning the Y-Z plane to the symmetry planes of objects \cite{chang2015shapenet,sun2018pix3d}, so the normal direction of the symmetry plane encodes most of the geometric information regarding the pose of the object.
To this end, we propose the \modelname{} network to detect the reflection symmetry from RGB images.
\emph{\modelname{}} combines the strength of learning-based recognition and geometry-based reconstruction methods.
It first enumerates the normal direction of the mirror plane from the image with a coarse-to-fine strategy and then verifies their correctness with a geometric-based neural network.
More specifically, we incorporate the concept of reflection symmetry into deep networks through plane-sweep cost volumes built from features of corresponding pixels, as shown in Figure \ref{fig:teaser}.
This allows us to accurately recover the normal direction of the mirror plane under the principle of shape-from-symmetry~\cite{HongW2004}.

\begin{figure*}[t]
  \centering
  \includegraphics[width=0.99\linewidth]{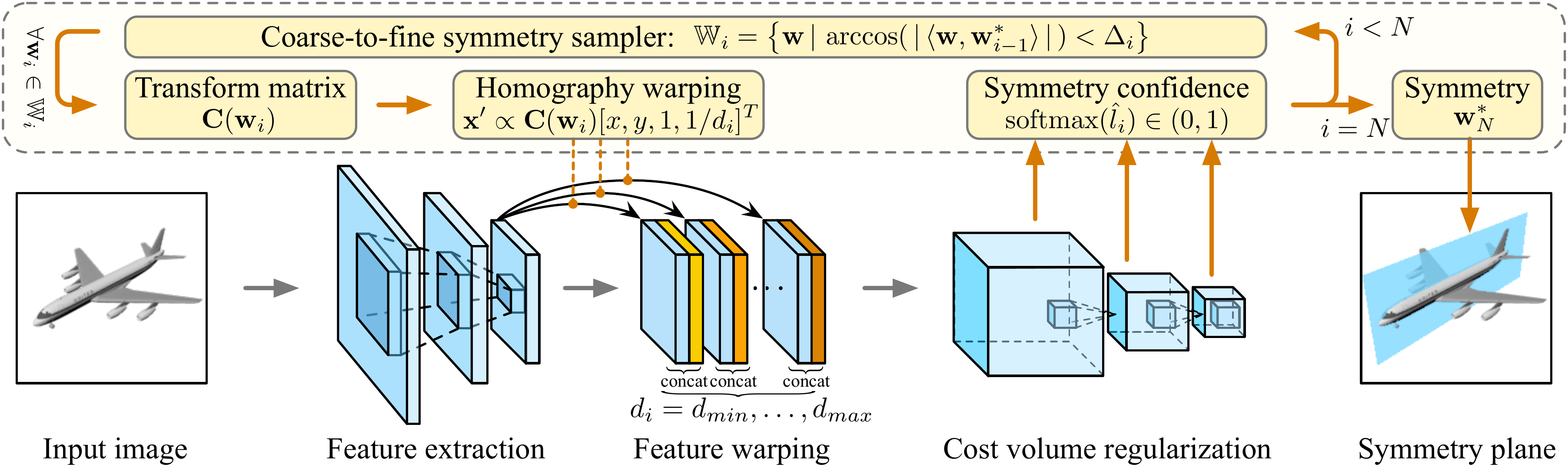}
  \caption{Overview of the \modelname{}. During inference, the coarse-to-fine symmetry sampler gives a list of candidate normal directions of the symmetry plane.  For each candidate symmetry plane,  a warping transformation matrix $\mathbf{C}$ is computed according to \Cref{eq:relationship}.  Input images first go through the feature extraction (backbone) network. Features are then warped by a warping module based on the symmetry transformation $\mathbf{C}$ and depth $d_i$. A cost volume is constructed by fusing the warped features and feeding into a 3D convolutional neural network for refinement.  The final confidence of each symmetry plane is predicted by aggregating the resulting depth probability tensor.
  }
  \label{fig:network}
\end{figure*}

The network (see Figure \ref{fig:network}) consists of a backbone feature extractor, a differentiable warping module for building the 3D cost volumes, and a cost volume network.
This framework naturally enables neural networks to utilize the information from corresponding pixels of reflection symmetry inside a single image.
We evaluate our method on the ShapeNet dataset \cite{chang2015shapenet} and Pix3D dataset \cite{sun2018pix3d}. Extensive comparisons and analysis show that by detecting and utilizing intra-image pixel correspondence from reflection symmetry, our method has better accuracy for recovering the normal direction of the symmetry plane and hence the object pose, even when the object is not perfectly symmetric.

Our main contributions are summarized as below:
\begin{itemize}[noitemsep,topsep=0pt,parsep=0pt,partopsep=0pt]
  \item we identify the problem of learning neural 3D reflection symmetry detector, in which the intra-image pixel correspondence of symmetry can be utilized for accurate plane normal estimation;
  \item we propose a novel framework that leverages single-view dense feature matching to estimate symmetry planes, significantly outperforming previous methods;
  \item we show that the learned symmetry planes benefit tremendously a variety of downstream tasks, including single-view pose recovery and depth estimation.
\end{itemize}

\begin{figure}[t]
  \centering
  \subfloat[][2D reflection symmetry \label{fig:symmetry:2d}]{
    \includegraphics[width=0.49\linewidth]{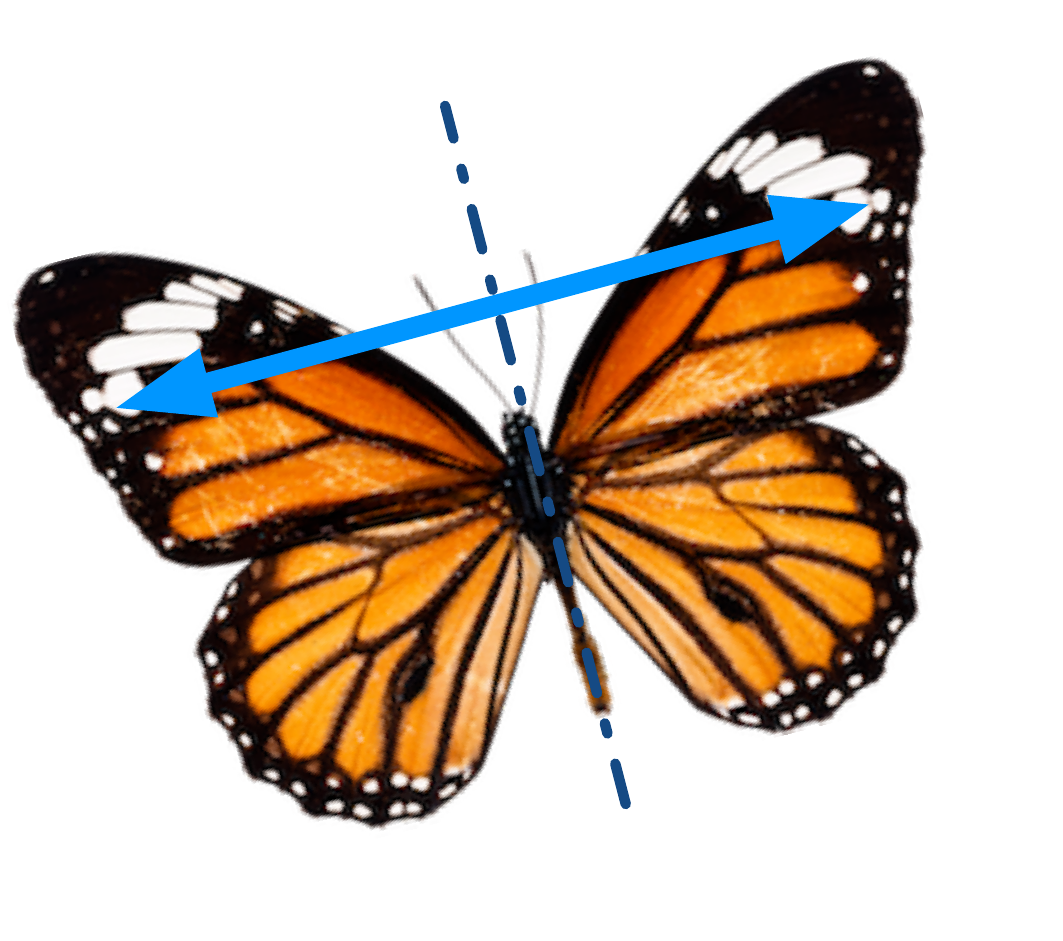}
  }
  \subfloat[][3D reflection symmetry \label{fig:symmetry:3d}]{
    \includegraphics[width=0.49\linewidth]{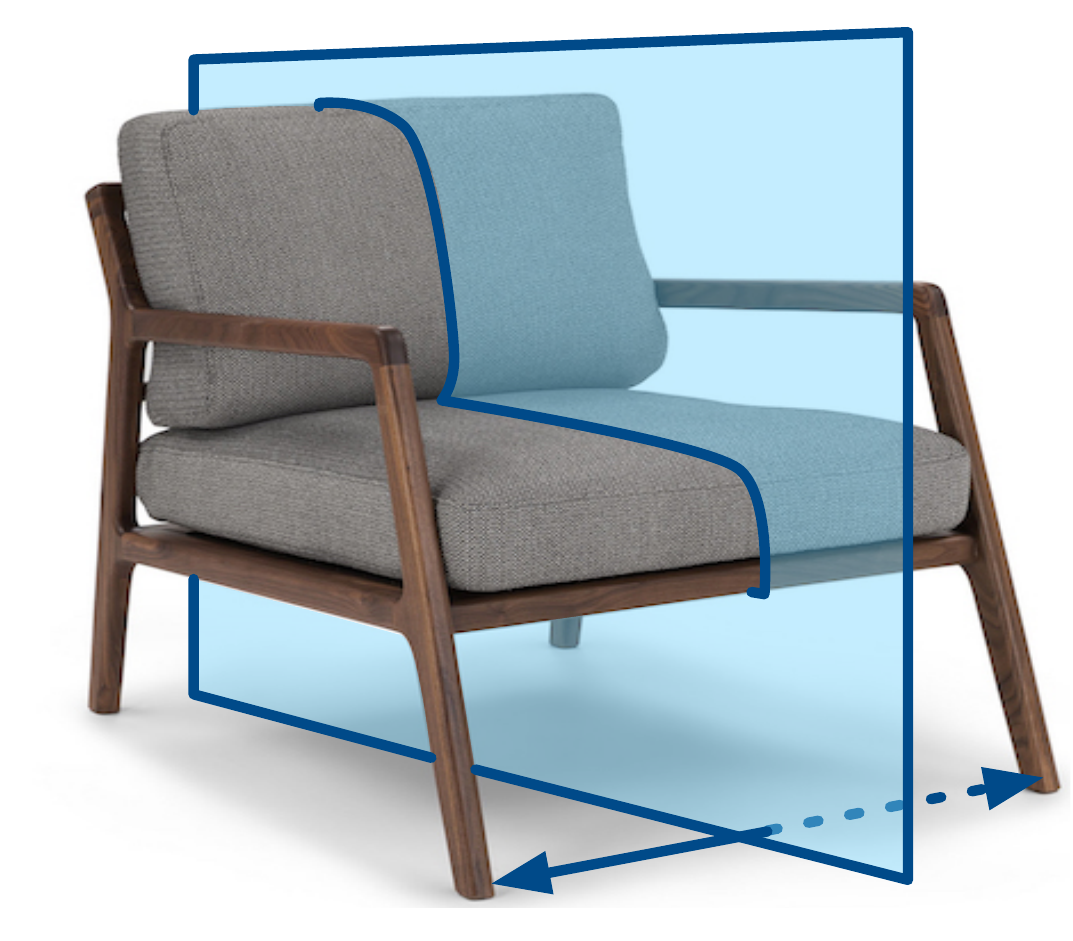}
  }
  \caption{Examples of 2D and 3D reflection symmetry reconstruction. 2D symmetries are not helpful for 3D understading due to lack of perspective distortion.}
  \label{fig:symmetry}
\end{figure}

\section{Related Work}

\paragraph{3D Reflection Symmetry.}
For many years, scientists from vision science and psychology have found that symmetry plays an important role in the human vision system \cite{Troje-symmetry,Vetter-symmetry2}.
People have exploited different kinds of symmetry for tasks such as texture impainting \cite{korah2008analysis}, unsupervised shape recovering \cite{wu2019unsupervised}, and image manipulation \cite{zhang2020portrait}.
Researchers have utilized the correspondences of symmetry to reconstruct shapes in different representations, such as points \cite{HongW2004}, curves \cite{HongW2004-ECCV}, and recent deep implicit fields \cite{xu2020ladybird}.
However, these methods either assume that the input camera pose or the symmetry plane is given or require its correspondence points.  This is because detecting 3D symmetry from a single view is challenging. 

\paragraph{Symmetry Detection.}
\cite{funk20172017} is a recent survey of existing 2D/3D symmetry detection methods.
On one hand, most of the geometry-based symmetry detection methods use handcrafted features and only work for 2D planar and front-facing objects \cite{loy2006detecting,ZabrodskyH1995-PAMI,KiryatiN1998-IJCV} as shown in \Cref{fig:symmetry:2d}.
The extracted 2D symmetry axes and correspondences cannot provide enough geometric cues for depth reconstruction.
In order to make reflection symmetry useful for depth reconstruction, it is necessary to detect the 3D mirror plane and corresponding points of symmetric objects (\Cref{fig:symmetry:3d}) from perspective images.
On the other hand, recent single-image processing neural networks \cite{chang2015shapenet,insafutdinov2018unsupervised,wang2019normalized,zhou2019continuity,yao2020front2back} can approximately recover the camera orientation with respect to the canonical pose, which gives a mirror plane of symmetry.
However, the camera poses from those data-driven networks are not accurate enough \cite{funk2017beyond}, because they cannot exploit the geometric constraints of symmetry.
To remedy the above issues, our \modelname{} tries to take the best of both worlds.
The proposed method first detects the 3D mirror plane of a symmetric object from an image and then recovers the depth map by finding the pixel-wise correspondence with respect to the symmetry plane, all of which are supported with geometric principles.
Our experiment (\Cref{sec:exp}) shows that \modelname{} is indeed much more accurate for 3D symmetry plane detection, compared to previous learning-based methods \cite{zhou2019continuity,xu2019disn}.

\paragraph{Learning-Based Single-Image 3D Understanding.}
Inspired by the success of CNNs in image classification and object detection,
multiple single-view learning-based 3D understanding tasks have been explored, including
depth estimation \cite{fu2018deep,chang2018pyramid}, camera pose recovery, etc.
Although these methods demonstrate promising results on benchmark datasets, the inferred results are not accurate enough for most subsequent 3D reconstruction purposes.
To alleviate this issue, our method leverages the symmetry prior by matching pixel-level features for accurate single-view 3D understanding.

\section{Methods}

\subsection{Camera Model and 3D Symmetry} \label{sec:symmetry}

Let $\object \subset \bbR^4$ be the set of points in the homogeneous coordinate that are on the surface of an object.  If we say $\object$ admits the \emph{symmetry}\footnote{An object might admit multiple symmetries. For example, a rectangle has two reflective symmetries and one rotational symmetry. We here only consider the principle symmetry.} with respect to a rigid transformation
$\M \in \bbR^{4 \times 4}$, it means that
\begin{equation}
    \forall \X \in \object: \M\X \in \object, \quad \mbox{and}\quad \feature(\X) = \feature(\M\X), \label{eq:sym}
\end{equation}
where $\X$ is homogeneous coordinates of a point on the surface of the object, $\M\X$ is the corresponding point of $\X$ with respect to the symmetry, and $\feature(\cdot)$ represents the surface properties at a given point, such as the surface material and texture.  For example, if an object has reflection symmetry with respect to the Y-Z plane in the world coordinate, then we have its transformation $\M_x = \mathrm{diag}(-1, 1, 1, 1)$.  \Cref{fig:symmetry} shows an example of 3D reflection symmetry.

Given two 3D points $\X,\X' \in \object$ in the homogeneous coordinate that are associated by the symmetry transform $\X' = \M \X$, their 2D projections $\x$ and $\x'$ must satisfy the following conditions:
\begin{equation}
  \x = \K\Rt\X/d,\quad \mbox{and} \quad  \x' = \K\Rt\X'/d'. \label{eq:proj2}
\end{equation}
Here, we keep all vectors in $\mathbb{R}^4$.  $\x=[x,y,1,1/d]^T$ and $\x'=[x',y',1,1/d']^T$ represent the 2D coordinates of the points in the pixel space, $d$ and $d'$ are the depth in the camera space, 
$\K \in \mathbb{R}^{4\times4}$ is the camera intrinsic matrix, and $
\Rt = \begin{bsmallmatrix}
  \R & \t \\
   0  & 1
\end{bsmallmatrix}
$ is the camera extrinsic matrix that rotates and translates the coordinate from the object space to the camera space.

From Equation \eqref{eq:proj2}, we can derive the following constraint for their 2D projections $\x$ and $\x'$:
\begin{equation}
  \x' \propto \underbrace{\K\Rt\M\Rt^{-1}\K^{-1}}_{\C} \x \doteq \C \x. \label{eq:correspondence}
\end{equation}
We use the proportional symbol here as the 3rd dimension of $\x'$ can always be renormalized to one so the scale factor does not matter. The constraint in \Cref{eq:correspondence} is valuable to us because the neural network now has a geometrically meaningful way to check whether the estimated depth $d$ is reasonable at $(x,y)$ by comparing the image appearance at $(x, y)$ and $(x', y')$, where $(x', y')$ is computed from \Cref{eq:correspondence} given $x$, $y$, and $d$.  If $d$ is a good estimation, the two corresponding image patches should be similar due to $\feature(\X) = \feature(\X')$ from the symmetry constraint in \Cref{eq:sym}. This is often called \emph{photo-consistency} in the literature of multi-view steropsis \cite{furukawa2009accurate}.

An alternative way to understand \Cref{eq:correspondence} is to substitute $\X' = \M \X$ into Equation \eqref{eq:proj2} and treat the later equation as the projection from another view. By doing that, we reduce the problem of shape-from-symmetry to two-view stereopsis, only that the stereo pair is in special positions.

\paragraph{Reflection Symmetry in 3D.}
\Cref{eq:correspondence} gives us a generalized way to represent any types of symmetry with matrix $\C=\K\Rt\M\Rt^{-1}\K^{-1}$.  For reflection symmetry, a more intuitive parametrization is to use the equation of the symmetry plane in the camera space.  Let $\tilde \x \in \mathbb{R}^3$ be the coordinate of a point on the symmetry plane in the camera space.  The equation of the symmetry plane can be written as
\begin{equation}
\w^T \tilde{\x} + 1 = 0,
\end{equation}
where we use $\w\in\mathbb{R}^3$ as the parameterization of symmetry.  The relationship between $\C$ and $\w$ is
\begin{equation}
  \C(\w) = \K \left(\mathbf{I} - \frac{2}{\|\w\|_2^2}\begin{bmatrix}
  \w \\
  \mathbf{0} \\
  \end{bmatrix}
  \begin{bmatrix} \w^T& \mathbf{1} \end{bmatrix} \right)\K^{-1}. \label{eq:relationship}
\end{equation}
We derive \Cref{eq:relationship} in the supplementary material.  The goal of reflection symmetry detection is to recover $\w$ from images.

On the first impression, one may wonder why $\Rt$ (i.e., camera poses) in \Cref{eq:correspondence} has 6 degrees of freedoms (DoFs) while $\w$ only has 3.  This is due to the specialty of reflection symmetry.  Rotating the camera with respect to the normal of the symmetry plane (1 DoF) and translating the camera along the symmetry plane (2 DoFs) cannot change the relative pose of the camera with respect to the symmetry plane.  Therefore the number of DoFs in reflection symmetry is indeed $6-1-2=3$.

\paragraph{Scale Ambiguity.}
\begin{figure}[t]
  \centering
  \includegraphics[width=0.99\linewidth]{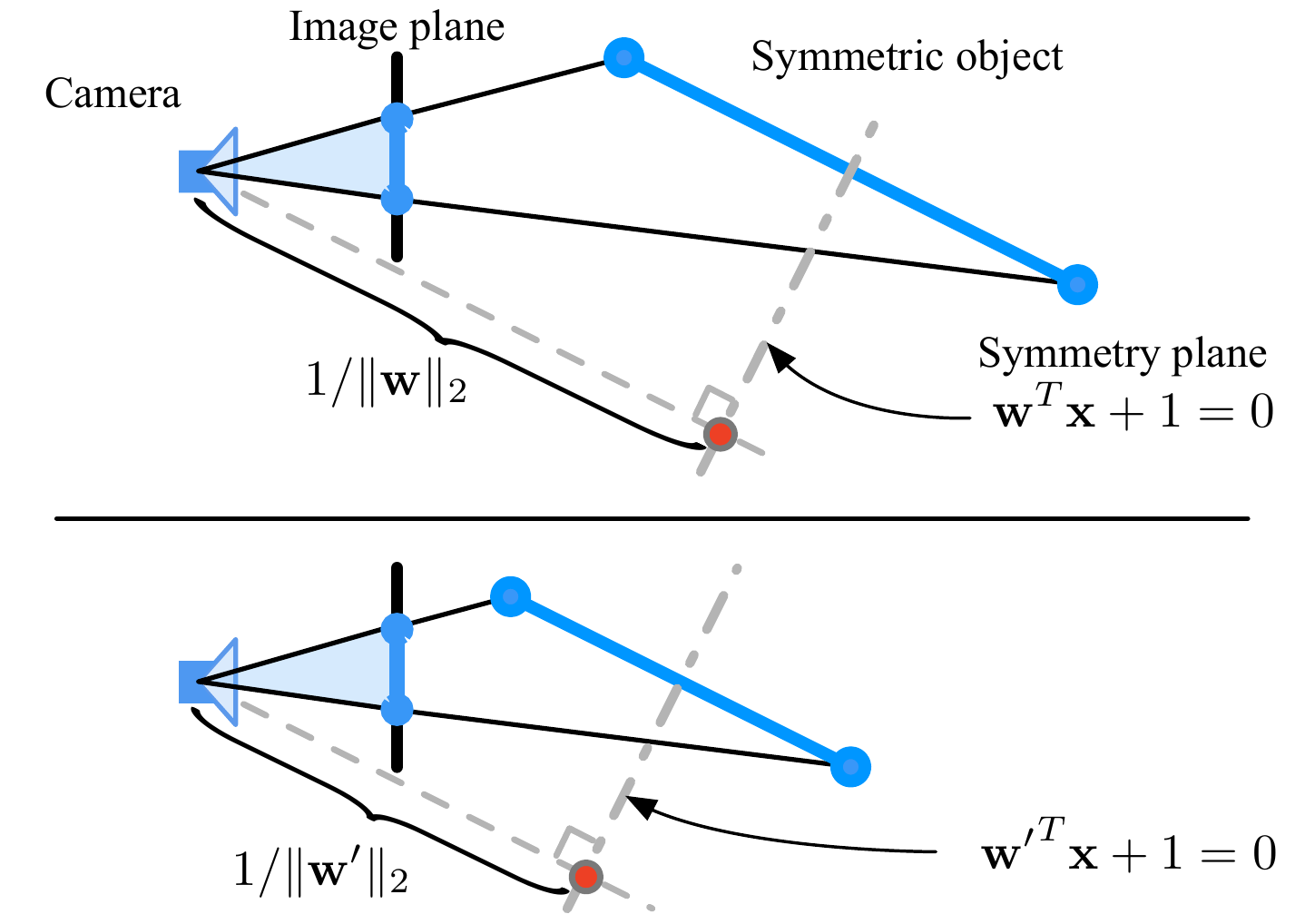}
  \caption{Illustration of scale ambiguity. We display two scenes that only differ by a scale $c$.  The images of the two scenes are exactly the same, but the distances between the origin and two symmetry planes are different, i.e., $\|\w\|_2=c\|\w'\|_2$.}
  \label{fig:scale-ambi}
\end{figure}
Similar to structure-from-motion in which it is impossible to determine the absolute size of scenes \cite{ma2012invitation}, shape-from-symmetry also has a scale ambiguity.
This is demonstrated in \Cref{fig:scale-ambi}.
In the case of reflection symmetry, we cannot determine the value of $\|\w\|_2$, i.e., the symmetry plane's distance from the origin, from a single image without relying on size priors, as it is always possible to scale the scene by a constant (and thus scale $\|\w\|_2$) without affecting images.
Therefore, we fix $\|\w\|_2$ to be a constant and leave the ambiguity as it is.
In other words, NeRD is designed only to recover the normal direction of the symmetry plane.
For real-world applications, this scale ambiguity can be resolved when the object size or the distance between the object and the camera is known.

\subsection{Overall Pipeline of \modelname{}} \label{sec:overall}

\paragraph{Motivation.}  \Cref{sec:symmetry} provides us a geometric way to verify whether a given $\w$ is valid: For each pixel $(x,y)$, we check if there exists a $d$ so that the image feature at $(x, y)$ and its mirror point $(x', y')$ are similar, where $(x', y')$ are computed with \Cref{eq:correspondence}.  If $\w$ is correct, then for pixels whose mirror parts are not occluded, we should be able to find their corresponding pixels that are similar to themselves.  To utilize such an idea, we turn the problem of regressing $\w$ into a classification problem: We first enumerate possible plane normal directions and use a neural network to verify whether these directions are closed to the real symmetry planes or not.

\begin{figure}[t]
  \centering
  \begin{minipage}[b]{.49\textwidth}
  \setlength{\lineskip}{0pt}
  \includegraphics[width=0.45\linewidth]{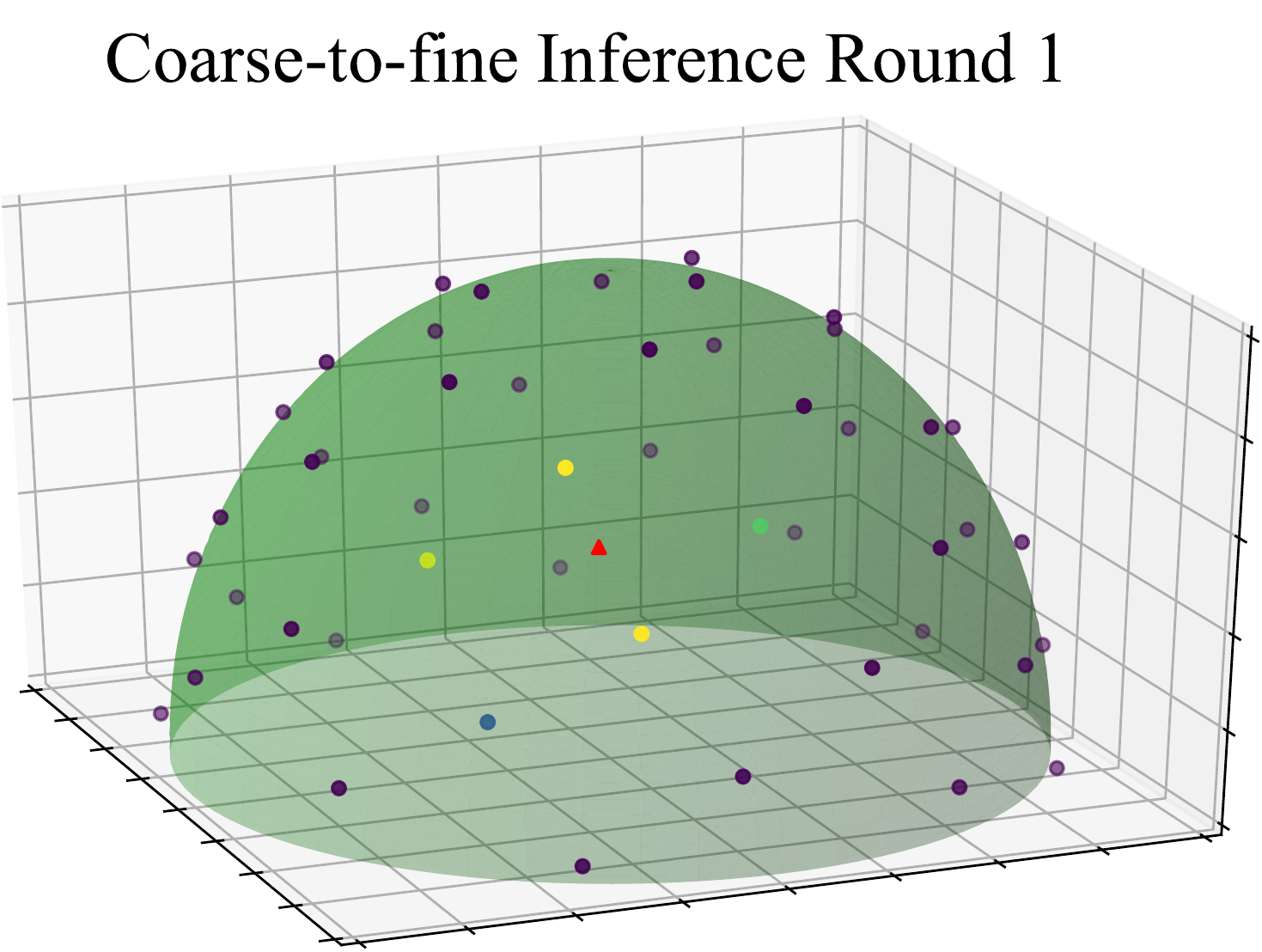}%
  \hspace{0.8em}
  \includegraphics[width=0.45\linewidth]{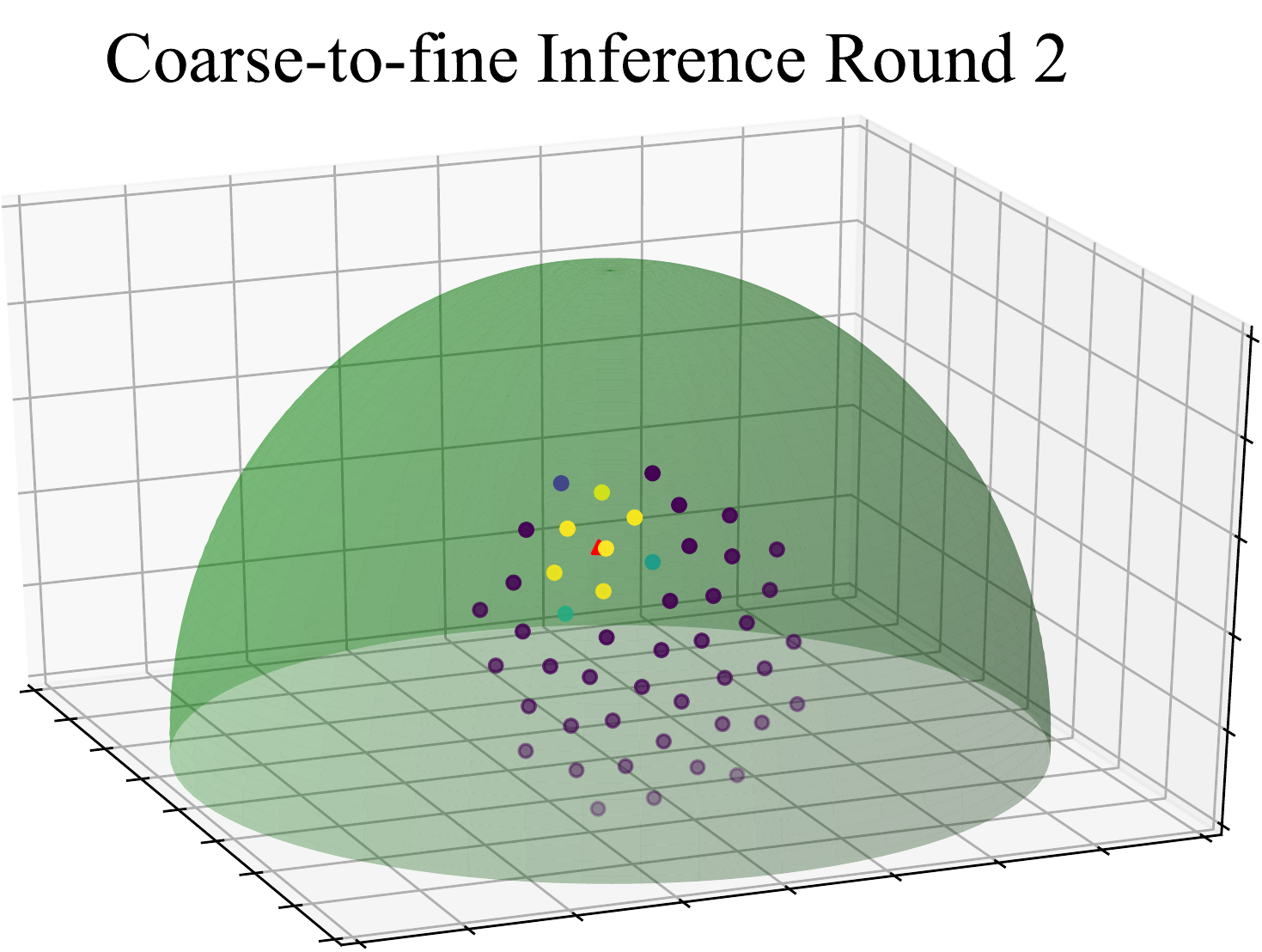}
  \\
  \includegraphics[width=0.45\linewidth]{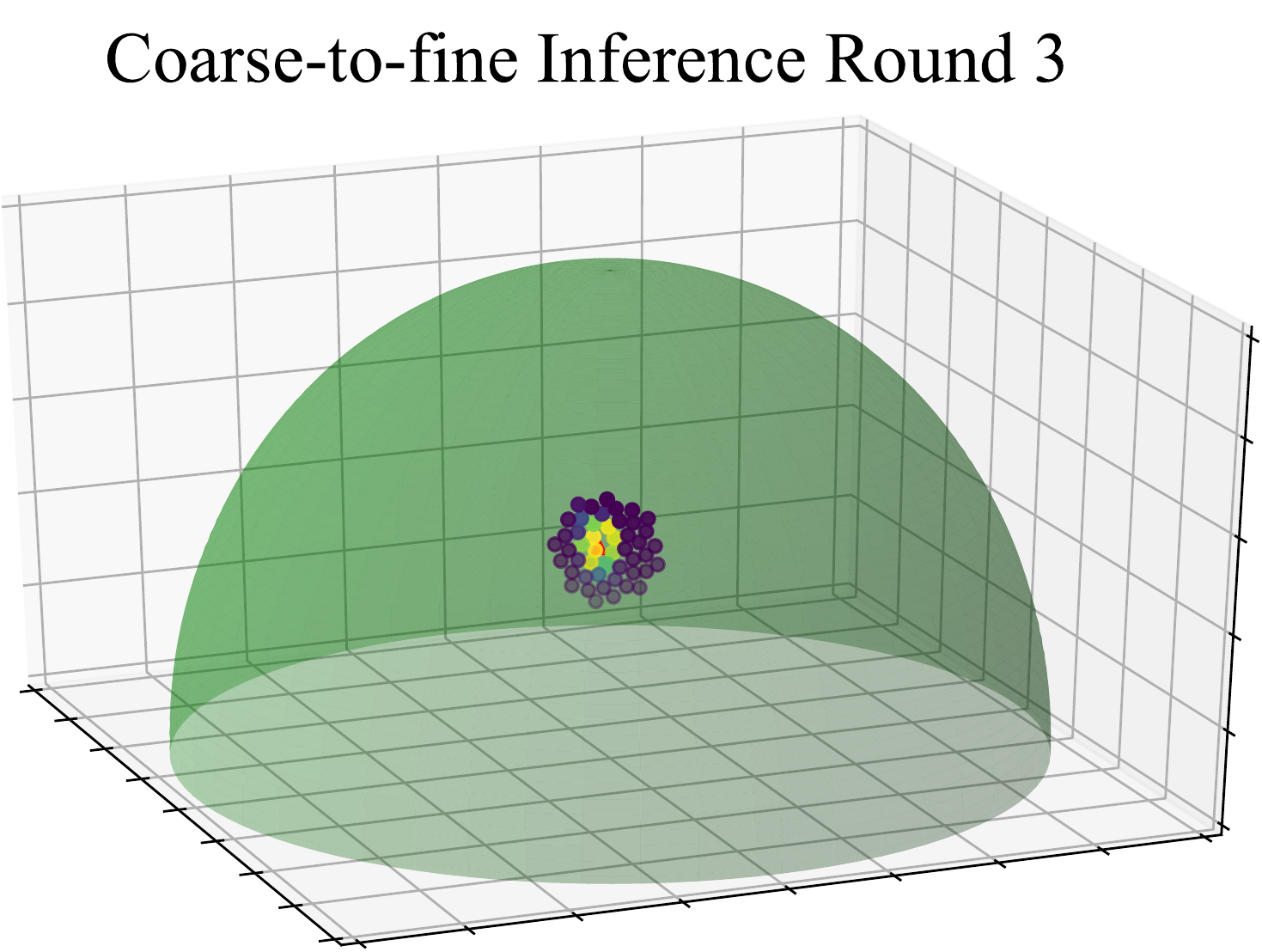}%
  \hspace{0.8em}
  \includegraphics[width=0.45\linewidth]{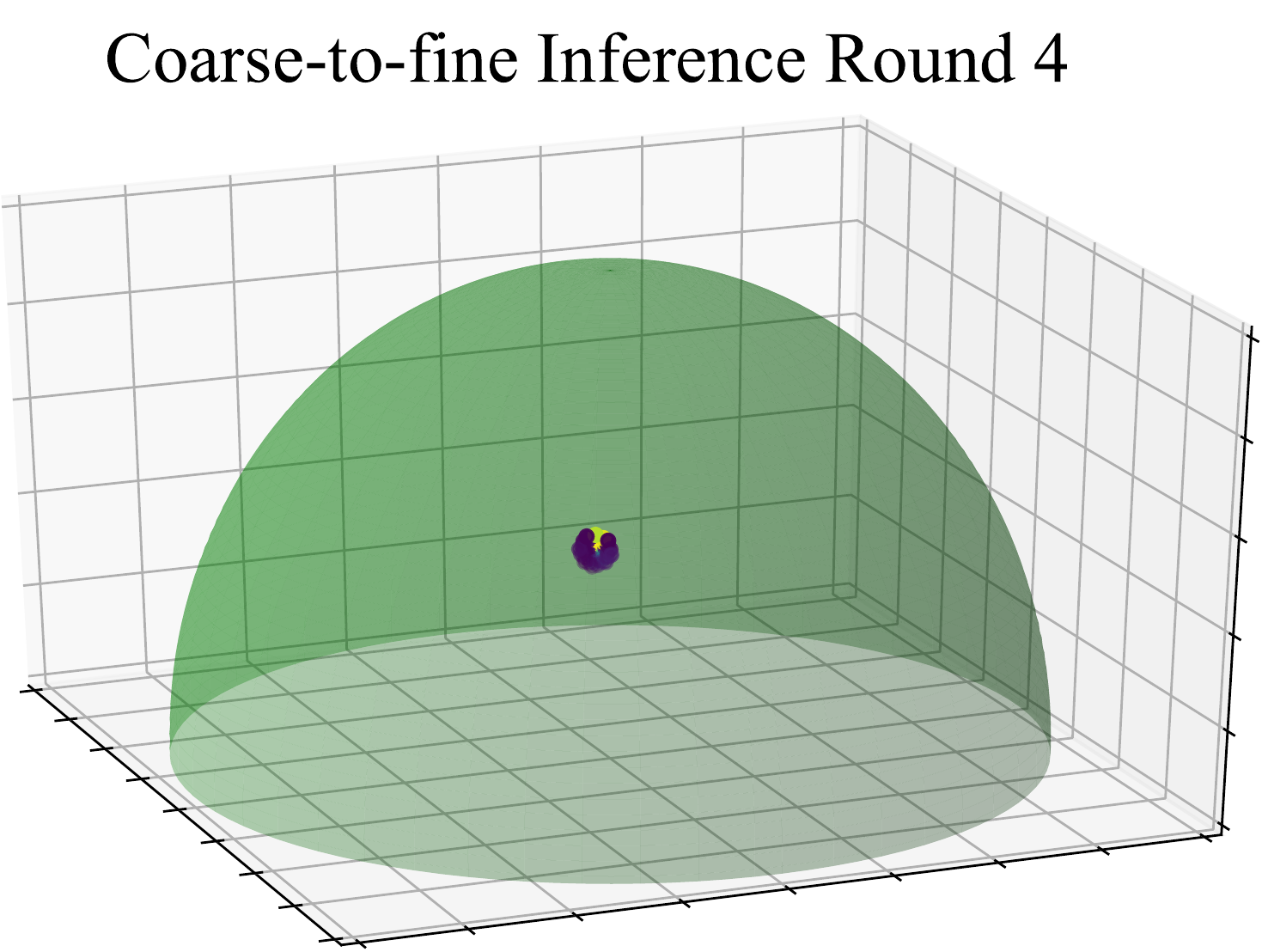}%
  \end{minipage}%
  \caption{Illustration of the process of coarse-to-fine inference. We show the sampled normal direction in a 4-round coarse-to-fine inference.  The color of points represents the scores from symmetry confidence network.}
  \label{fig:demo}
\end{figure}

\paragraph{Methods.}
\Cref{fig:network} illustrates the overall pipeline of \modelname{} during inference.
For each input image, we compute its 2D feature map (\Cref{sec:backbone}) and generate a list of candidate normal directions of its symmetry plane.
For each candidate normal $\w$, we use it to warp the 2D feature map and construct an initial 3D cost volume (\Cref{sec:warping}) for photo-consistency matching.
After that, the cost volume network (\Cref{sec:costvolume}) converts the cost volume tensor into a confidence value.
We pick $\w$ with the highest confidence as the resulting normal direction of the symmetry plane.

A brute-force enumeration of $\w$ is slow, especially when high precision is needed.
To accelerate it, \modelname{} uses a coarse-to-fine strategy, which we will describe in detail in \Cref{sec:sampler}.
\Cref{fig:demo} illustrates the process of coarse-to-fine inference.
In $i$th round of inference, the coarse-to-fine sampler samples $N$ candidates symmetry plane $\{\w_{i}^k\}_{k=1}^{K}$ uniformly and evaluate their confidence with our neural network.
Then, we find the pose $\w_i^{*}$ with the highest confidence score and limit the symmetry sampler to the nearby region around it.
This process is repeated until we achieve the desired accuracy.

\subsection{Backbone Network}\label{sec:backbone}
The goal of the {\em backbone network} is to extract 2D features from images.  We use a modified ResNet-like network as our backbone.  To reduce the memory footprint, we first down-sample the image with a stride-2 $5 \times 5$ convolution.  After that, the network has 8 \emph{basic blocks} \cite{he2016deep} with ReLU activation.  The 5th basic block uses stride-2 convolution to further downsample the feature maps.   The number of channels is 64.  The output feature map $\mathbf{F}$ has dimension $\lfloor \frac H 4\rfloor \times \lfloor \frac W 4\rfloor \times 64$.  The network structure diagram is shown in the supplementary materials. %

\subsection{Feature Warping Module} \label{sec:warping}

The function of the {\em feature warping module} is to construct the initial 3D cost volume tensor $\V(x,y,d)$ for photo-consistency matching.  We discretize $d$ so that $d \in \depthSet=\{d_{\min} + \frac{i}{D-1}(d_{\max}-d_{\min}) \mid i = 0,1,\dots,D-1\}$ to make the 3D cost volume homogeneous to 3D convolution, in which $d_{\min}$ and $d_{\max}$ is the minimal and maximal depth we want to predict and $D$ is the number of sampling points for depth.
As mentioned in \Cref{sec:symmetry}, the correctness of $d$ at $(x, y)$ correlates with the appearance similarity of the image patch at pixels represented by $\x$ and $\C\x$.  Therefore, we set $\V$ by concatenating the backbone features at these two locations, i.e.,
\begin{equation}
    \V(x, y, d) = \Big[\mathbf{F}(x, y), \,\mathbf{F}(x', y')\Big],
\end{equation}
where $[\,x', y', 1, 1/d'\,]^T  \propto \C [\,x, y, 1, 1/d\,]^T$, i.e., $(x', y')$ being the projection of the mirror point of the pixel $(x, y)$ assuming its depth is $d$.
Here $\mathbf{F}$ is the backbone feature, and $\C$ is computed from the sampled symmetry plane $\hat\w$.  We apply bilinear interpolation to access the features at non-integer coordinates.  The dimension of the cost volume tensor is $\lfloor \frac H 4\rfloor \times \lfloor \frac W 4\rfloor \times D \times 32$.

\subsection{Cost Volume Network} \label{sec:costvolume}
The goal of the cost volume network is to turn the initial 3D cost volume tensor $\V$ from the feature warping module into a confidence value representing whether the current pose $\w$ is close to the ground truth.  It may also predict a depth probability tensor $\P(x,y, d):=\Pr[\D(x,y)=d]$ for downstream tasks (\Cref{sec:depth}).  The cost volume network uses matrix multiplication on the channel dimension to check for the photo-consistency on $\V$.  However, the initial cost volume aggregated from image features can be noisy.  Thus, we use a network consists of multiple 3D convolution layers that are capable of regularizing the cost volume information. We aggregate the multi-resolution encoder features with max-pool operators and then apply the sigmoid function to normalize the confidence values into $[0, 1]$.

\subsection{Symmetry Sampler} \label{sec:sampler}
\paragraph{Inference.} As shown in \Cref{fig:demo}, the symmetry sampler uniformly samples $\{\w_{i}^k\}_{k=1}^{K}$ from $\mathbb{W}_i \subset \mathbb{R}^3$ using the Fibonacci lattice \cite{zhou2019neurvps,gonzalez2010measurement}, where $\mathbb{W}_i$ is the sampling space of the $i$th round of inference. In the first round, candidates are sampled from the surface of a unit hemisphere.  For the following rounds, we set $\mathbb{W}_i=\{\w\in\mathbb{S}^2\,|\,\arccos(|\langle\w,\w_{i-1}^*\rangle|) < \Delta_i\}$ to be a spherical cap, where $\w_{i-1}^*$ is the optimal $\w$ from the previous round and $\Delta_i$ is a hyper-parameter.

\paragraph{Training.}
During training, we sample symmetry planes for each image according to the hyper-parameter $\Delta_i$.
For the $i$th level,  symmetry candidates are sampled from $\{\hat\w\in\mathbb{S}^2\,|\,\arccos(|\langle\w,\hat\w\rangle|) \le \Delta_i\}$,
where $\w$ is the ground truth symmetry pose.
We also add a random sample $\hat{\w} \in \mathbb{S}^2$ to reduce the sampling bias.
For each sampled $\hat\w$, its confidence labels is $l_{i}=\mathbf{1}[\arccos(|\langle\w,\hat\w\rangle|) < \Delta_{i}]$ for the $i$th level.
The training error could be written as
\begin{align*}
  L_{\mathrm{cls}} &= \sum_{i}\mathrm{BCE}(\hat l_{i}, l_{i}),
\end{align*}
where $\mathrm{BCE}$ represents the binary cross entropy error, and $\hat l_i$ is predicted confidence of $\hat{w}$ for the $i$th level in the coarse-to-fine inference.

\subsection{Applications}\label{sec:depth}   
In this section, we introduce some potential applications of reflection symmetry detection that benefit from the accurate normal direction of the reflection symmetry plane.

\paragraph{Pose Recovery.} In the problem of pose recovery, the goal is to find the pose of an object from an RGB image, in which people normally set up the canonical space of objects so that objects are symmetric with respect to the X-Z plane or the Y-Z plane \cite{chang2015shapenet}.
Because \modelname{} is able to pinpoint the normal direction of the symmetry plane, we can accurately determine 2 DoFs of the 6 DoFs pose with our geometry-based method.  For the rest 4 DoFs, we can still resort to data-driven approaches (e.g., direct regression) with neural networks.

\paragraph{Depth Estimation.}  As we construct cost volumes (i.e., depth probability tensors) in the symmetry detection pipeline (\Cref{sec:costvolume}), it is straightforward to use it for a geometry-based depth estimation.   With the estimated $\w^*$, we compute the expectation of depth from the probability tensor $\P$ as the depth map prediction $\hat\D$.  This is sometimes referred as \emph{soft argmin} \cite{kendall2017end}.  Mathematically, we have
\begin{align}
  \hat\D(x,y) &= \frac{1}{|\depthSet|} \sum_{d \in \depthSet} d\P(x,y, d). \label{eq:softargmin}
\end{align}
We rescale the ground truth depth according to $\|\hat{\w}\|_2$ and add an additional $\ell_1$ term to the training loss as the supervision of depth:
\begin{align}
  L_{\mathrm{dpt}} &= \frac{1}{n}\sum_{x,y} \left|\hat\D(x,y) - \D(x,y)\right|,
\end{align}
where $n$ is the number of pixels.

\section{Experiments} \label{sec:exp}

\subsection{Datasets}
We conduct experiments on the synthetic ShapeNet dataset \cite{chang2015shapenet} and real-world Pix3D dataset \cite{sun2018pix3d}, in which models have already been processed so that in their canonical poses the Y-Z plane is the plane of the reflection symmetry.

\paragraph{ShapeNet.} We use the same camera pose, intrinsic, and train/validation/test split from a 13-category subset of the dataset as in R2N2 and others \cite{kar2017learning,wang2018pixel2mesh,choy20163d} to make the comparison easy and fair.
We exclude the lamp category as it contains many asymmetric objects.
We use Blender to render the images with resolution $256 \times 256$.

\paragraph{Pix3D.}
Pix3D \cite{sun2018pix3d} is a real-world dataset containing image-shape pairs with 2D-3D registrations.
To demonstrate the versatility of \modelname{}, we test \modelname{} on the Pix3D dataset.
We assume that the bounding boxes of objects have been detected, and we use them to crop the images for removing the background while maintaining the aspect ratio.
We rescale the resulting images to $256 \times 256$ and adjust the camera intrinsic matrix $\K$ accordingly and reject images extraordinary with focal lengths and depth values.
We randomly split the remaining data into train and test sets, which contain 5285 and 588 images, respectively.

\subsection{Implementation Details}
We implement \modelname{} in PyTorch.
We use the plane $x=0$ in the object space as the ground truth symmetry plane because it is explicitly aligned for each model by authors of ShapeNet.
We set $d_{\min}$ and $d_{\max}$ according to the depth distribution of the dataset, and use $D=64$ for the depth of the cost volume.
We use $N=4$ rounds in the coarse-to-fine inference, in each of which $K=32$ normal directions are sampled.
We choose $\Delta=[20.7^\circ, 6.44^\circ, 1.99^\circ, 0.61^\circ]$ according to the gap between near directions on the Fibonacci lattice.
Our experiments are conducted on two NVIDIA RTX 2080Ti GPUs.  We use Adam \cite{kingma2014adam} for training.
The learning rate is set to $3 \times 10^{-4}$ and batch size is set to 16 per GPU.
We train the \modelname{} for 40 epochs and decay the learning rate by a factor of 10 at the 30th epoch.
The overall inference speed is about 1 image per second per GPU.

\begin{table}[t]
\setlength{\tabcolsep}{1.2mm}
\renewcommand{\arraystretch}{1.15}
\resizebox{\linewidth}{!}{%
\begin{tabular}{|c|c|c|ccc|cccc|}
\hline
  & \multirow{2}{*}{\begin{tabular}[c]{@{}c@{}}backbone\\ (sec \ref{sec:backbone})\end{tabular}} & \multirow{2}{*}{\begin{tabular}[c]{@{}c@{}}cost volume\\ (sec \ref{sec:costvolume})\end{tabular}} & \multicolumn{3}{c|}{feature warping}        & \multicolumn{4}{c|}{error metrics}                            \\ \cline{4-10} 
  &                                                                                              &                                                                                                   & var          & avg          & cat          & avg          & med          & $\!<\!1^\circ$ & $\!<\!2^\circ$ \\ \hline
  a &                                                                                              & $\checkmark$                                                                                      &              &              & $\checkmark$ & $7.12^\circ$ & $\bm{0.54^\circ}$ & $66.8\%$       & $77.2\%$       \\
b & $\checkmark$                                                                                 &                                                                                                   &              &              & $\checkmark$ & $6.82^\circ$ & $0.99^\circ$ & $50.1\%$       & $70.1\%$       \\
c & $\checkmark$                                                                                 & $\checkmark$                                                                                      & $\checkmark$ &              &              & $6.33^\circ$ & $0.57^\circ$ & $68.1\%$       & $81.5\%$       \\
d & $\checkmark$                                                                                 & $\checkmark$                                                                                      &              & $\checkmark$ &              & $6.41^\circ$ & $0.66^\circ$ & $63.7\%$       & $77.7\%$       \\ \hline
  \textbf{e} & $\checkmark$                                                                                 & $\checkmark$                                                                                      &              &              & $\checkmark$ & $\bm{5.41^\circ}$ & $0.56^\circ$ & $\bm{68.2\%}$       & $\bm{81.5\%}$       \\ \hline
\end{tabular}%
}
\caption{Ablation study of 3D reflection symmetry detection on ShapeNet.}
  \label{tab:ablation}
\end{table}

\paragraph{Metrics.}
To better understand the performance of symmetry detection, we show two forms of metrics.  We plot a performance curve for each detector-dataset pair, in which the x-axis represents the angle accuracy and the y-axis represents the proportion of the data whose error is less than that.  We also report quantitative metrics, including the median and mean of the angle difference, and the percentages of testing images whose error is smaller than $0.5^\circ$, $1.0^\circ$, $2.0^\circ$, and $4.0^\circ$,  for the ease of comparison.

\subsection{Ablation Studies}
\begin{figure}[t]
    \newcommand{\lengthb}{0.49\linewidth}
    \centering
    \vspace{-2.3ex}
    \subfloat[][Curves on ShapeNet (synthetic)\label{fig:curve:shapenet}]{
      \includegraphics[width=\lengthb]{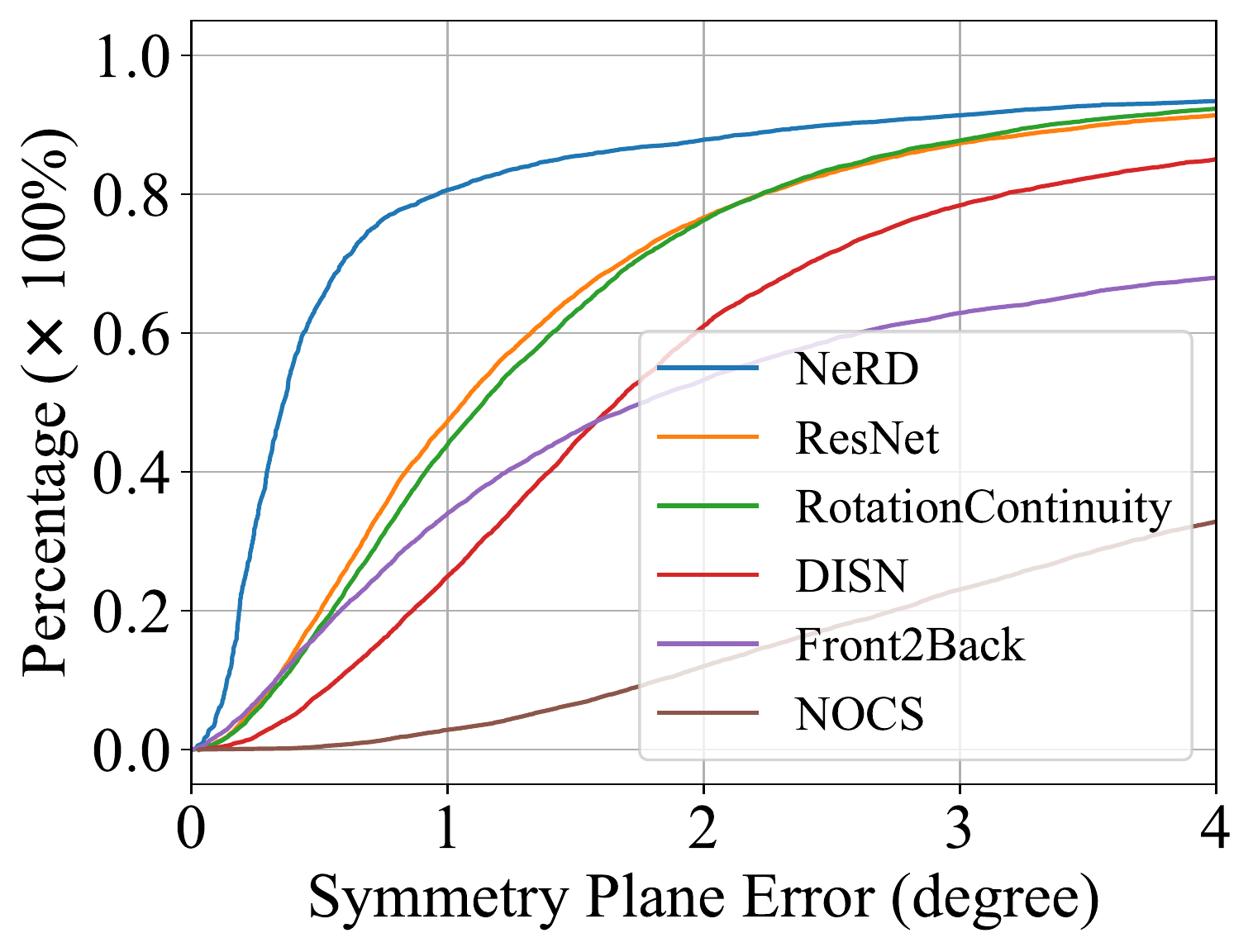}
    }%
    \subfloat[][Curves on Pix3D (real-world)\label{fig:curve:pix3d}]{
      \includegraphics[width=\lengthb]{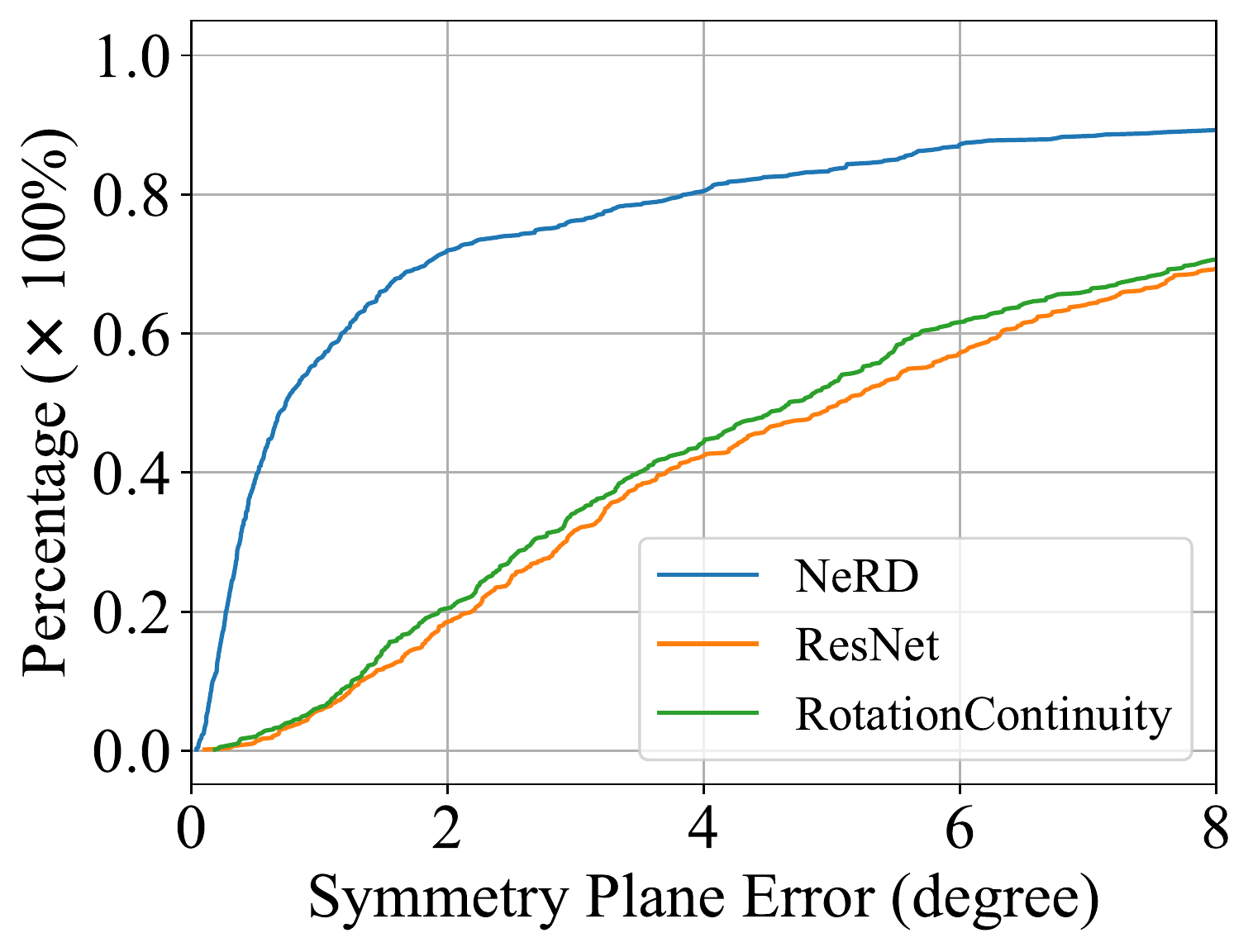}
    }%
    \caption{Performance curves of symmetry detection and camera pose recovery networks.  \emph{Higher is better.}}
    \label{fig:benchmark}
\end{figure}

\begin{table}[t]
\setlength{\tabcolsep}{2.0mm}
\renewcommand{\arraystretch}{1.2}
\resizebox{\linewidth}{!}{%
\begin{tabular}{|l|cc|ccc|}
\hline
                                             & avg               & med                & $\!<\!0.5^\circ$ & $\!<\!1.0^\circ$ & $\!<\!2.0^\circ$ \\ \hline
DISN \cite{xu2019disn}                       & $2.80^\circ$      & $1.65^\circ$       & $7.96\%$         & $24.9\%$         & $61.0\%$         \\
ResNet \cite{he2016deep}                     & $2.08^\circ$      & $1.06^\circ$       & $19.7\%$         & $47.3\%$         & $76.6\%$         \\
RotationContinuity \cite{zhou2019continuity} & $1.94^\circ$      & $1.14^\circ$       & $17.6\%$         & $43.9\%$         & $76.2\%$         \\
Front2Back \cite{yao2020front2back}          & $9.41^\circ$      & $1.76^\circ$       & $16.8\%$         & $34.0\%$         & $53.2\%$         \\
NOCS~\cite{wang2019normalized}               & $9.95^\circ$      & $6.18^\circ$       & $0.39\%$         & $2.83\%$         & $11.9\%$         \\
\textbf{NeRD}                                & $\bm{1.58^\circ}$ & $\bm{0.36^\circ}$  & $\bm{64.5\%}$    & $\bm{80.6\%}$    & $\bm{87.8\%}$    \\ \hline
\end{tabular}%
}
\caption{
Performance of symmetry detection and object pose recovery algorithms on ShapeNet.
We report the normal direction error of the predicted symmetry planes.
We note that NOCS \cite{wang2019normalized} requires ground truth object shapes as input.
}
  \label{tab:results:shapenet}
\end{table}

\begin{table}[t]
\setlength{\tabcolsep}{2.0mm}
\renewcommand{\arraystretch}{1.2}
\resizebox{\linewidth}{!}{%
\begin{tabular}{|l|cc|ccc|}
\hline
                                             & avg               & med                 & $\!<\!1.0^\circ$ & $\!<\!2.0^\circ$ & $\!<\!4.0^\circ$ \\ \hline
ResNet \cite{he2016deep}                     & $8.01^\circ$      & $5.06^\circ$        & $5.78\%$         & $18.5\%$         & $42.3\%$         \\
RotationContinuity \cite{zhou2019continuity} & $7.91^\circ$      & $4.67^\circ$        & $6.12\%$         & $20.4\%$         & $44.3\%$         \\
\textbf{NeRD}                                & $\bm{3.37^\circ}$ & $\bm{0.73^\circ}$   & $\bm{56.3\%}$    & $\bm{71.9\%}$    & $\bm{80.4\%}$    \\ \hline
\end{tabular}%
}
\caption{
  Performance of symmetry detection and object pose recovery algorithms on real-world dataset Pix3D \cite{sun2018pix3d}.
  We report the normal direction error of the predicted symmetry planes.
}
  \label{tab:results:pix3d}
\end{table}

We conduct ablation studies to justify each component in \modelname{}. 
In \Cref{tab:ablation}, we analyize the function of three main components of \modelname{}: the 2D backbone network (\Cref{sec:backbone}), feature warping module (\Cref{sec:warping}), and the cost volume network (\Cref{sec:costvolume}).
The second column of \Cref{tab:ablation} represents whether we use the feature from the 2D backbone or just RGB values with a single $1 \times 1$ convolution to construct the cost volume.
Comparing (a) and (e), we find that removing the 2D backbone degrades the performance, especially at the region $>2^\circ$.
We think this is because the 2D backbone network increases the receptive field, which makes our method more robust.
The third column represents whether we want to replace the cost volume network with a simple max-pool layer.
Results in (b) and (e) show that the cost volume network is the key component for an accurate symmetry detector.
Finally, we study the different pooling schemes in the feature warping module.
From (c), (d), and (e), we find that the feature concatenation and variance pooling gives the best results, while the average pooling performs poorly in the high-precision region ($<1^\circ$).
This matches our intuition in \Cref{sec:symmetry} that \modelname{} compares the feature to check photo-consistency.

\subsection{Symmetry Detection on Synthetic Datasets}
\vspace{-0.6ex}
\paragraph{Baselines.}
We briefly introduce some state-of-the-art single-view symmetry detection and pose estimation baselines.
Probably the plainest way to estimate the 3D symmetry plane $\w$ is direct regression \cite{funk2017beyond}.
We implement it with ResNet-50 \cite{he2016deep} and train it with L1 loss.
RotationContinuity~\cite{zhou2019continuity} identifies a 6D representation of rotation which they claim is more suitable for learning.
We also implement it and train with L1 loss.
DISN~\cite{xu2019disn} also implements its 6D representation for ShapeNet but is trained with L2 loss. We report the performance of their pre-trained model.
Front2Back~\cite{yao2020front2back} is a recent work that detects the 3D symmetry plane, which first predicts a depth map and then fits the symmetry plane with a traditional method~\cite{mitra2006partial}.  We report the performance of their results provided by the authors.
NOCS~\cite{wang2019normalized} predicts a coordinate of normalized object coordinate space for each pixel and recovers the pose with Umeyama algorithm \cite{umeyama1991least}.
Following their paper, we train the NOCS estimator on ShapeNet and use their code to recover the orientation of objects from prediction.

\begin{table}[t]

\centering
\setlength{\tabcolsep}{1.5mm}
\renewcommand{\arraystretch}{1.2}
\resizebox{\linewidth}{!}{%
\begin{tabular}{|c|cccc|ccc|}
\hline
                                       & absRel & sqRel  & rmse  & mae   & $\!<\!\delta^1$ & $\!<\!\delta^2$ & $\!<\!\delta^3$ \\ \hline
  DORN \cite{fu2018deep}               & $0.028$  & $0.0014$ & $0.026$ & $0.020$ & $30.8\%$          & $54.1\%$          & $69.0\%$          \\
  GeoNet \cite{yin2018geonet}             & $0.028$  & $0.0013$ & $0.025$ & $0.019$ & $29.7\%$          & $53.4\%$          & $69.2\%$          \\
  Hourglass \cite{newell2016stacked}   & $0.026$  & $0.0012$ & $0.024$ & $0.018$ & $33.0\%$          & $56.9\%$          & $71.5\%$          \\
  DenseDepth \cite{alhashim2018high}   & $0.024$  & $0.0011$ & $0.022$ & $0.017$ & $36.3\%$          & $60.5\%$          & $74.6\%$          \\
  Pixel2Mesh \cite{wang2018pixel2mesh} & $0.102$  & $0.0546$ & $0.032$ & $0.073$ & $28.6\%$          & $49.2\%$          & $62.3\%$          \\
  DISN \cite{xu2019disn}               & $0.040$  & $0.0030$ & $0.038$ & $0.028$ & $24.0\%$          & $43.4\%$          & $57.8\%$          \\
  \textbf{\modelname{}}                & $\bm{0.019}$ & $\bm{0.0009}$ & $\bm{0.021}$ & $\bm{0.011}$ & $\bm{49.5\%}$ & $\bm{71.9\%}$  & $\bm{82.3\%}$ \\ \hline
  \modelname{}*                        & $0.015$  & $0.0006$ & $0.018$ & $0.011$ & $60.2\%$          & $78.7\%$          & $86.5\%$          \\ \hline
\end{tabular}%
}
  \caption{
    Quantitative comparison of \modelname{} and other baseline methods on ShapeNet.  We set $\delta = 1.01$.
    \modelname{}* uses the ground truth symmetry plane as input.
}
\label{tab:comparison}

\end{table}

\begin{figure}[t]
    \centering
    \includegraphics[width=\linewidth]{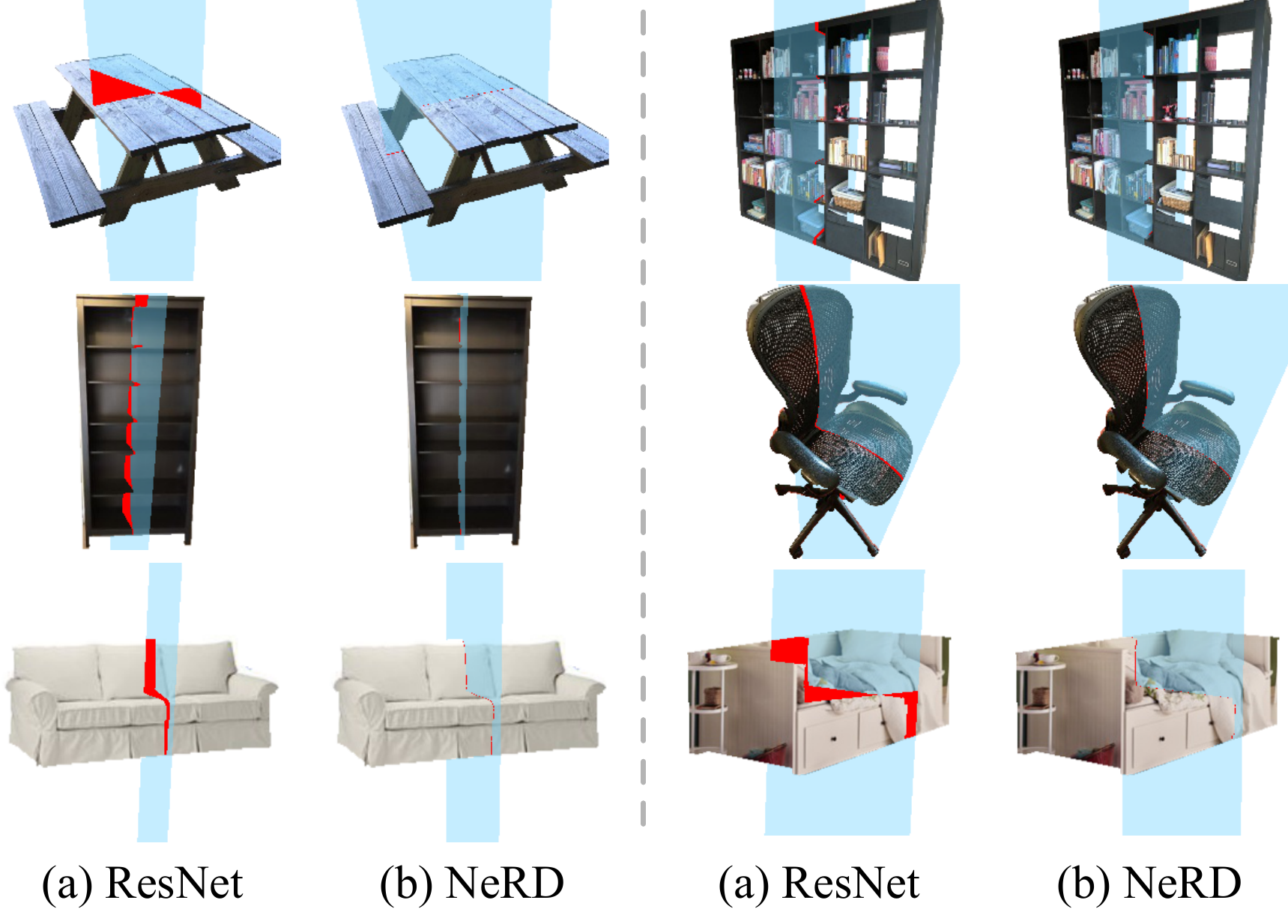}
    \caption{Qualitative results on the task of symmetry detection on Pix3D.  We show the detected symmetry planes from ResNet and our \modelname{}.  Errors of symmetry planes (pixels between the predicted and ground truth planes) are \textcolor{red}{highlighted}.}
    \label{fig:result:pix3d}
\end{figure}
\begin{figure*}[t]
    \centering
    \includegraphics[width=\linewidth]{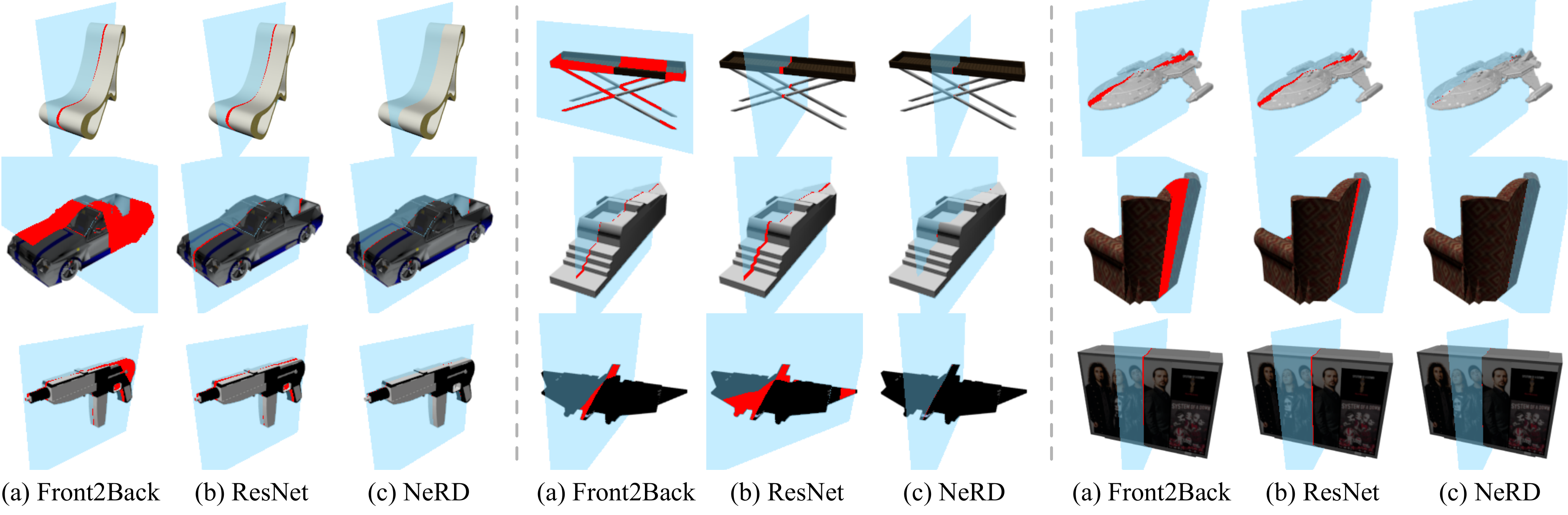}
    \caption{Qualitative results on ShapeNet.  Errors of symmetry planes (pixels between the predicted and ground truth planes) are \textcolor{red}{highlighted}.}
    \label{fig:result:shapenet}
\end{figure*}

\begin{figure*}[tp]
    \centering
    \includegraphics[width=\linewidth]{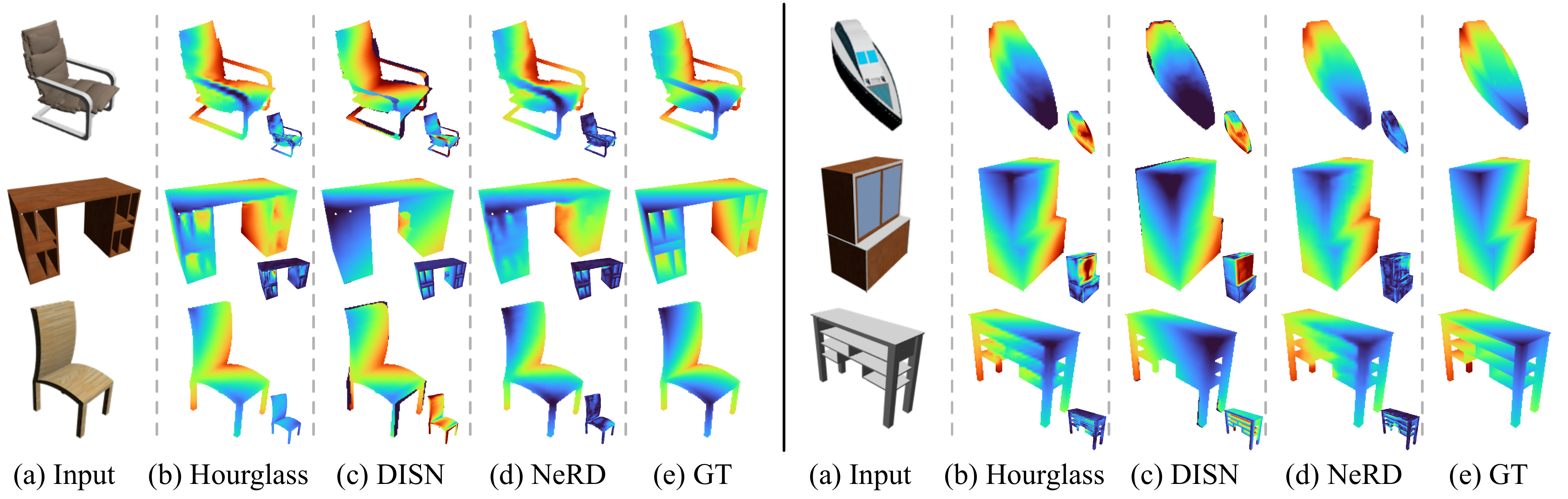}
    \caption{Qualitative results on the task of of depth estimation.  We visualize the depth maps from Pixel2Mesh \cite{wang2018pixel2mesh}, DISN \cite{xu2019disn}, and our \modelname{} on ShapeNet.  The per-pixel errors are plotted at the lower right corner.  \textcolor{blue}{Bluish} color represents smaller values for error and depth.}
    \label{fig:visualization}
\end{figure*}

\paragraph{Results.}
\Cref{tab:results:shapenet} and \Cref{fig:curve:shapenet} show the comparison on ShapeNet.
By utilizing geometric cues from symmetry, our approach significantly outperforms previous state-of-the-art methods.
The performance gap is larger in the region of higher precision ($<1^\circ$).
For example, \modelname{} can achieve an accuracy of $0.5^\circ$ on about $70\%$ of testing cases, while direct regression with ResNet and other baselines can only reach that on less than $20\%$ of data.
Such phenomena indicate that the intra-image correspondence does help algorithms recover symmetry planes more accurately, while naive CNNs can only roughly predict the plane normal by interpolating from training data.
We also find that end-to-end approaches that directly predict the symmetry plane (ResNet, DISN, NeRD, etc) performs better than the methods which require heavier post-processing (NOCS and Front2Back).
This hints us that using a loss function that is more directly related to the goal has an advantage.

\subsection{Symmetry Detection on Real-World Datasets}
\Cref{tab:results:pix3d} and \Cref{fig:curve:pix3d} show the comparison on the real-world Pix3D dataset.
\modelname{} outperforms the naive CNN regression, and the margin is even bigger compared to the results on ShapeNet.
We hypothesize that this is because images in Pix3D use a larger number of camera configurations, including different focal lengths and object positions with respect to the focal center, while the dataset has fewer images. This requires more generalizability from the algorithms.
Our geometry-based approach shines here because it can rely on the cues from correspondence to find the symmetry planes.
Also, it is hard for naive convolutional neural networks to make use of the camera intrinsics, which varies from images to images, unlike ShapeNet.
In contrast, \modelname{} uses camera intrinsic matrices in the feature warping module (\Cref{sec:warping}) and thus generalizes better when dealing with different camera configurations.

\subsection{Depth Estimation as an Application}
As mentioned in \Cref{sec:depth}, \modelname{} can be modified as a symmetry-guided depth estimator.  We compare it with popular monocular depth estimation networks~\cite{fu2018deep,yin2018geonet,newell2016stacked,alhashim2018high} and shape reconstruction networks~\cite{wang2018pixel2mesh,xu2019disn}.
The results on the task of \emph{depth estimation} are shown in \Cref{tab:comparison}.
\modelname{} outperforms both monocular depth estimation networks and shape reconstruction networks.
Besides, \modelname{}*, the variant of \modelname{} that uses the ground truth symmetry plane instead of the one predicted in coarse-to-fine inference, only slightly outperforms the standard \modelname{}.
These behaviors indicate that detecting symmetry planes and incorporating photo-consistency priors of reflection symmetry into the neural network makes the task of single-view reconstruction less ill-posed and thus can improve the performance.

\subsection{Visualization}
We visualize our results in  \Cref{fig:result:pix3d} and \Cref{fig:result:shapenet}. 
We have the following observations:
1) our method outperforms previous methods on unusual objects, e.g. chairs in atypical shapes.
This indicates that previous learning-based methods need to extrapolate from seen patterns and cannot generalize to unusual images well, while our method relies more on geometry cues from symmetry, a more reliable source of information for 3D understanding.
2) \modelname{} gives accurate symmetry planes even on challenging camera poses such as the orientation from the back of chairs.
We believe that this is because geometric information from correspondence helps to pinpoint the normal of symmetry planes.

In \Cref{fig:visualization}, we show sampled results of depth maps.
Visually, \modelname{} gives the most accurate results among all the tested methods.
For example, it can capture the details of desk frames and the shapes of ship cabins. %
Results from the hourglass network are also sharp but are less accurate, which may be the sign of overfitting.
In the region such as the chair armrests and table legs, \modelname{} can recover the depth more accurate compared to the baseline methods.
This is because, for \modelname{}, pixel-matching based on photo-consistency in those areas is easy and can provide a strong signal.%

\section*{Acknowledgement}
This work is supported by the research grant from Sony, the ONR grant N00014-20-1-2002, and the joint Simons Foundation-NSF DMS grant 2031899.  We also thank Li Yi from Google Research for his comments.

{
\small
\bibliographystyle{ieee_fullname}
\bibliography{paper}
}

\clearpage
\appendix
\section{Supplementary Materials}
\subsection{Derivation of $\C(\w)$}
Let $\tilde\x_0 \in \mathbb{R}^3$ be a point in the camera space and $\tilde\x_1$ be its mirror point with respect to the symmetry plane
\begin{equation}
    \w^T \tilde \x + 1 = 0.
\end{equation}
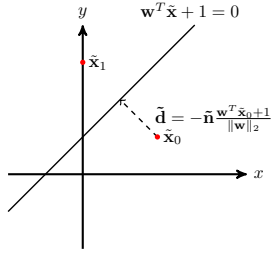
\begin{wrapfigure}{r}{0.44\linewidth}
\resizebox{\linewidth}{!}{%
\begin{tikzpicture}[
    scale=3,
    axis/.style={very thick, ->, >=stealth'},
    important line/.style={thick},
    every node/.style={color=black}
    ]
    \draw[axis] (-0.5,0)  -- (1.1,0) node(xline)[right] {$x$};
    \draw[axis] (0,-0.5) -- (0,1.0) node(yline)[above] {$y$};
    \draw[important line] (.-0.5,-0.25) coordinate (A) -- (.75,1)
        coordinate (B) node[above] {$\w^T\tilde\x+1=0\;\;$};
    \coordinate (x0) at (.50,.25); 
    \coordinate (x1) at (.00,.75); 
    \coordinate (xm) at (.25,.50); 
    \draw[important line,dashed,->] (x0) -- node[right] {$\;\;\mathbf{\vec{d}}=-\mathbf{\vec{n}} \frac{\w^T\tilde\x_0+1}{\|\w\|_2}$} (xm);
    \fill[red] (x0) circle (.5pt) node[right] {$\tilde\x_0$};
    \fill[red] (x1) circle (.5pt) node[right] {$\tilde\x_1$};
\end{tikzpicture}
}
\caption{Illustration of reflection symmetry with two points.}
\label{fig:mirroring}
\end{wrapfigure}

\Cref{fig:mirroring} illustrates the process of miring a point in 2D, where the red dots are the pair of points $\tilde\x_0$ and $\tilde\x_1$ the line in the middle is the symmetry plane whose normal $\mathbf{\vec{n}}=\frac{\w}{\|\w\|_2}$.
The distance between $\tilde\x_0$ and the symmetry plane is 
$\frac{\w^T\tilde\x_0+1}{\|\w\|_2}$, according to the formula of distance from a point to a plane.  Therefore, we have
\begin{equation}
    \tilde\x_1 = \tilde\x_0 - 2\frac{\w^T\tilde\x_0+1}{\|\w\|_2^2}\w.
\end{equation}
We could also write this in matrix form:
\begin{equation}
    \begin{bmatrix} \tilde\x_1 \\ 1 \end{bmatrix} =
    \begin{bmatrix} \tilde\x_0 \\ 1 \end{bmatrix} \begin{bmatrix}
      \mathbf{I} - \frac{2\w\w^T}{\|\w\|_2^2} & -\frac{2\w}{\|\w\|_2^2} \\
      \mathbf{0} & 1 \\
    \end{bmatrix}.
\end{equation}

\begin{figure}[t]
\centering
\includegraphics[width=.95\linewidth]{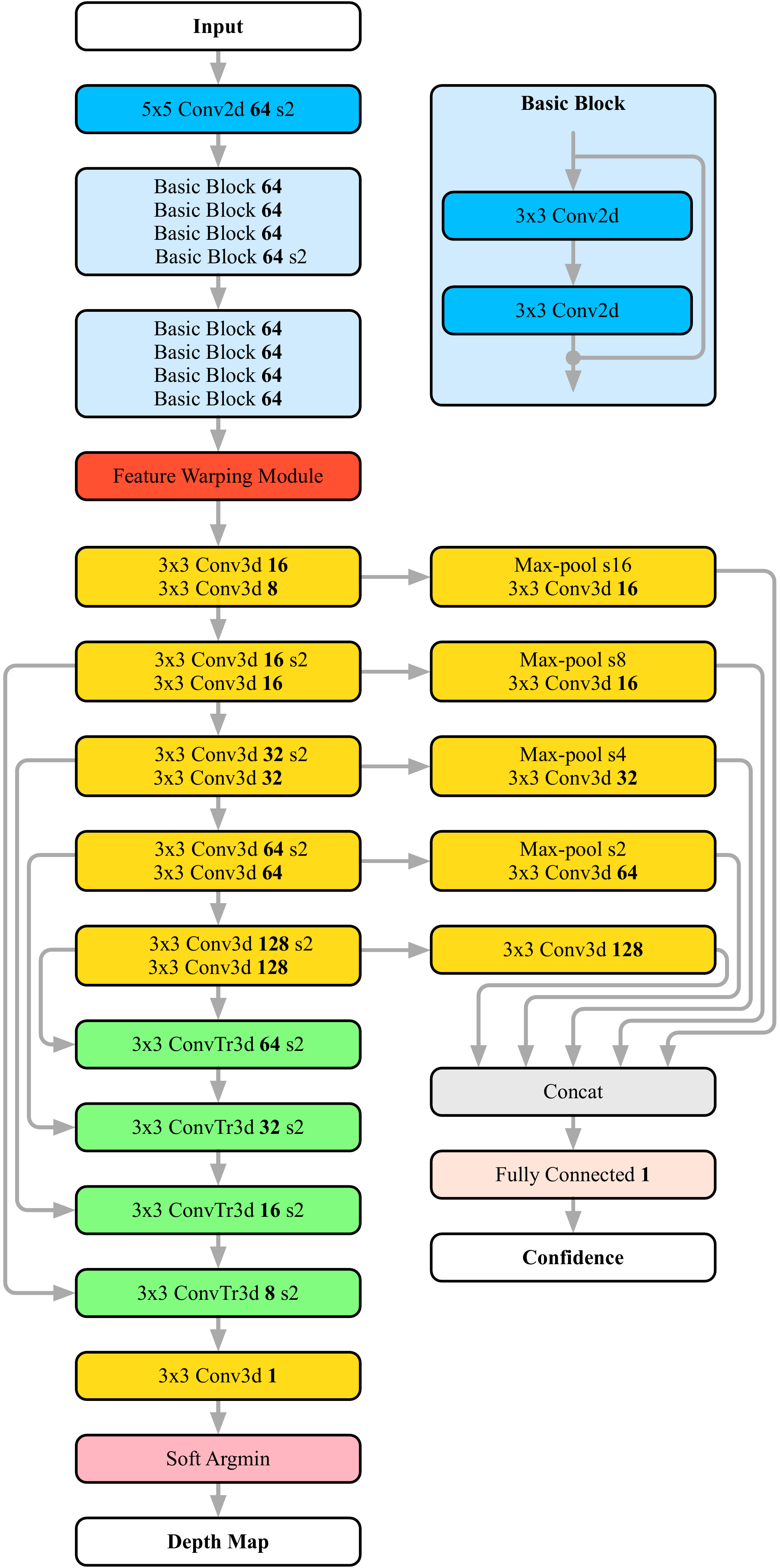}
\caption{Illustration of \modelname{}'s network architecture.  We show the resulting number of channels after each operator in \textbf{bold}. ``s2'' represents stride-2 operators.}
\label{fig:architecture}
\end{figure}

Because the transformation between the camera space and the pixel space is given by
\begin{equation}
    \x = \K \begin{bmatrix} \tilde\x \\ 1 \end{bmatrix},
\end{equation}
we finally have
\begin{align*}
  \C(\w) &= \K \begin{bmatrix}
  \mathbf{I} - \frac{2\w\w^T}{\|\w\|_2^2} & -\frac{2\w}{\|\w\|_2^2} \\
  \mathbf{0} & 1 \\
  \end{bmatrix}\K^{-1} \\
  &= \K \left(\mathbf{I} - \frac{2}{\|\w\|_2^2}\begin{bmatrix}
  \w \\
  \mathbf{0} \\
  \end{bmatrix}
  \begin{bmatrix} \w^T& \mathbf{1} \end{bmatrix} \right)\K^{-1}.
\end{align*}

\subsection{Network Architecture}

We display the \modelname{}'s network architecture in \Cref{fig:architecture}.

\subsection{Illustration of Coarse-to-Fine Inference}
\begin{figure}[t]
  \centering
  \setlength{\lineskip}{0pt}
  \def\mywidth{0.245\linewidth}

\includegraphics[width=\mywidth]{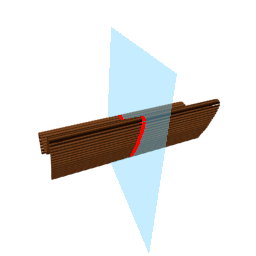}%
\includegraphics[width=\mywidth]{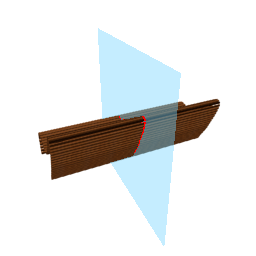}%
\includegraphics[width=\mywidth]{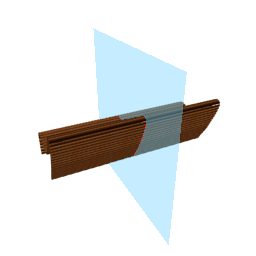}%
\includegraphics[width=\mywidth]{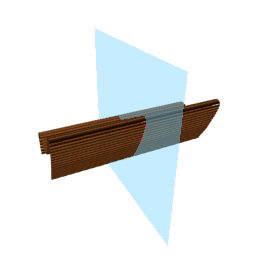}%

\includegraphics[width=\mywidth]{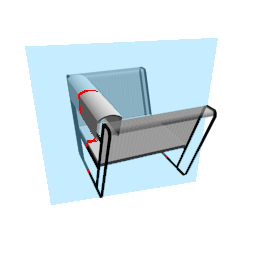}%
\includegraphics[width=\mywidth]{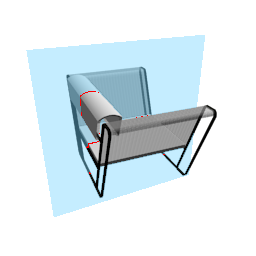}%
\includegraphics[width=\mywidth]{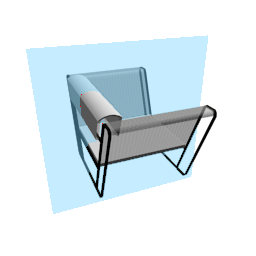}%
\includegraphics[width=\mywidth]{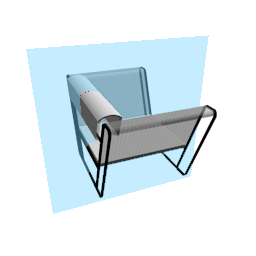}%

\includegraphics[width=\mywidth]{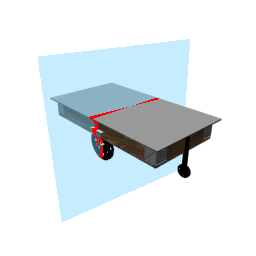}%
\includegraphics[width=\mywidth]{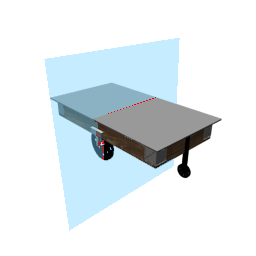}%
\includegraphics[width=\mywidth]{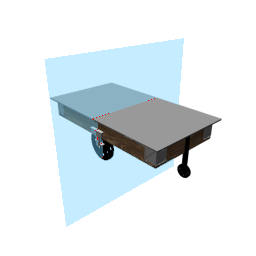}%
\includegraphics[width=\mywidth]{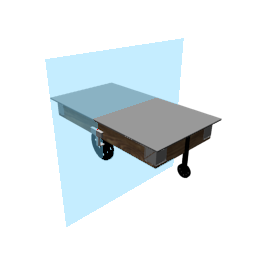}%

\includegraphics[width=\mywidth]{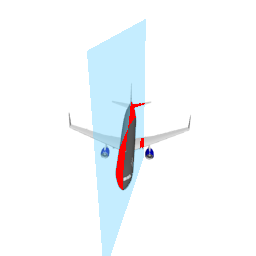}%
\includegraphics[width=\mywidth]{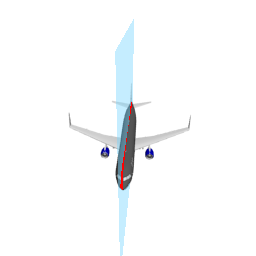}%
\includegraphics[width=\mywidth]{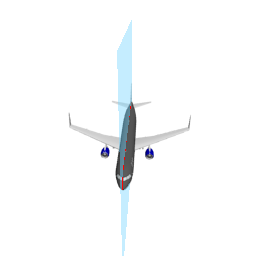}%
\includegraphics[width=\mywidth]{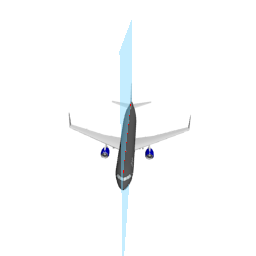}%

\includegraphics[width=\mywidth]{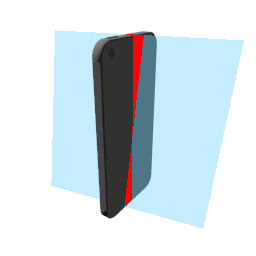}%
\includegraphics[width=\mywidth]{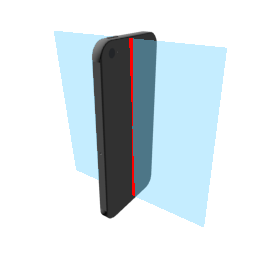}%
\includegraphics[width=\mywidth]{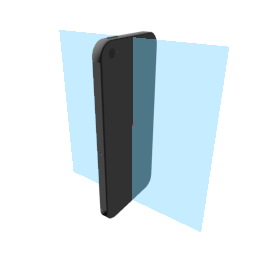}%
\includegraphics[width=\mywidth]{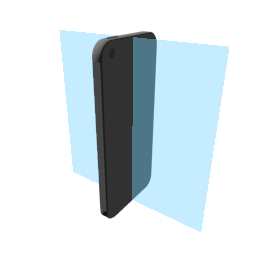}%

  \vspace{5pt}
\small
  \begin{minipage}[t]{\mywidth}\centering (a) Round \\$i=1$\end{minipage}%
  \begin{minipage}[t]{\mywidth}\centering (b) Round \\$i=2$\end{minipage}%
  \begin{minipage}[t]{\mywidth}\centering (c) Round \\$i=3$\end{minipage}%
  \begin{minipage}[t]{\mywidth}\centering (d) Round \\$i=4$\end{minipage}%
  \vspace{10pt}

  \caption{
    Illustration of the coarse-to-fine inference on sampled images from ShapeNet.
    The symmetry plane with the highest confidence score in each round of coarse-to-fine inference is drawn.
    In the $i$th round, we determine the normal of symmetry plane to the accuracy of $\Delta_i$, where $\Delta=[20.7^\circ, 6.44^\circ, 1.99^\circ, 0.61^\circ]$ are set according to the gap between nearly directions from the number of direction samples per round $K=32$.
  }
  \label{fig:supp:c2f}
  \vspace{10pt}
\end{figure}

\Cref{fig:supp:c2f} shows the process of coarse-to-fine inference on sampled images from ShapeNet.
We display the symmetry plane with the highest confidence score in each round of coarse-to-fine inference.
In the $i$th round, we determine the normal of symmetry plane to the accuracy of $\Delta_i$, where $\Delta=[20.7^\circ, 6.44^\circ, 1.99^\circ, 0.61^\circ]$ are set according to the gap between near directions from the number of direction samples per round $K=32$.
The coarse-to-fine inference dramatically reduces the number of samples required to achieve a certain level of accuracy.
As shown in the figure, the precision of the predicted plane increases with the number of rounds in the coarse-to-fine inference.

\subsection{Failure Cases}
\begin{figure}[tbp]
  \centering
  \setlength{\lineskip}{0pt}
  \includegraphics[width=0.66\linewidth]{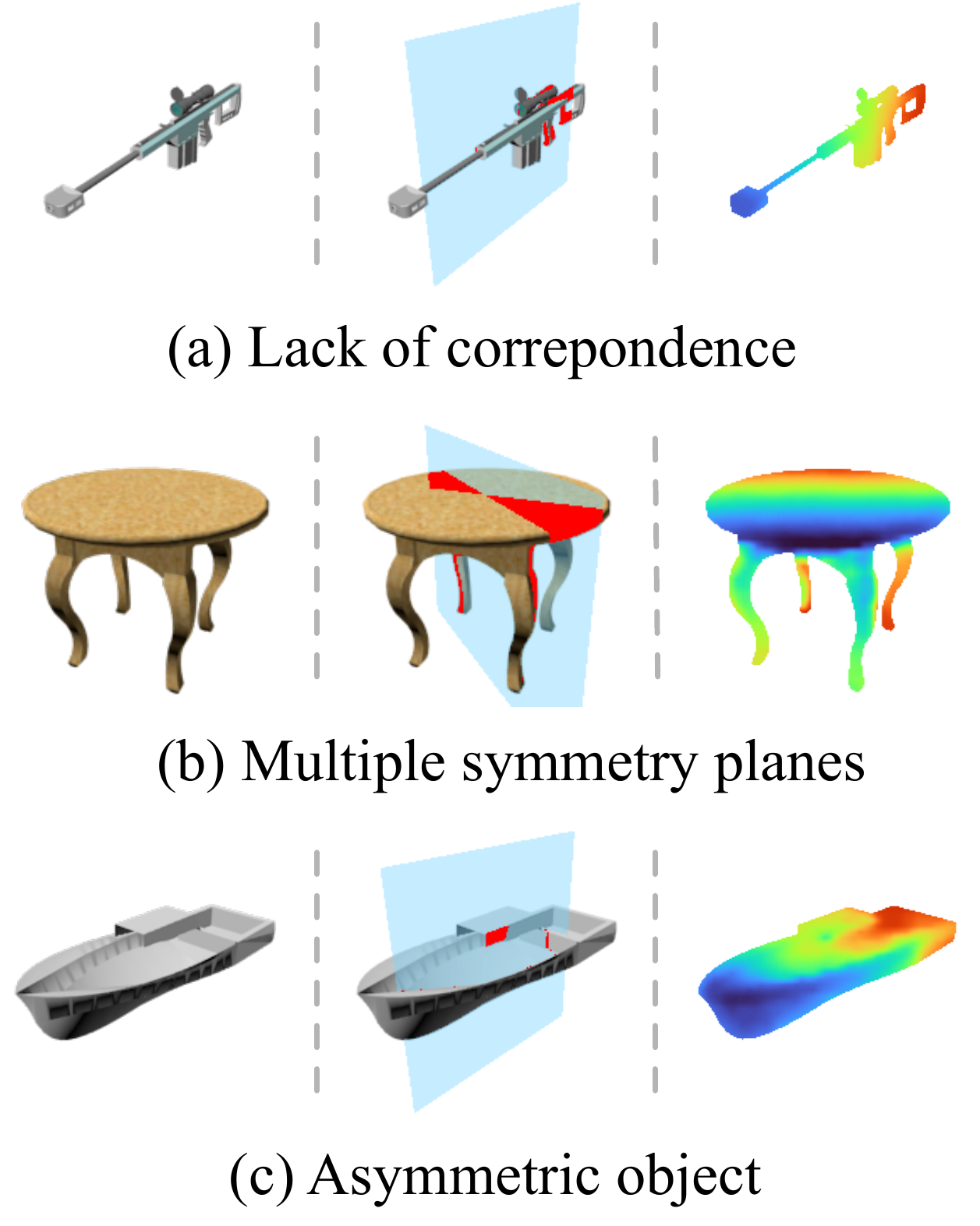}
  \caption{Sampled failure cases of \modelname{} on ShapeNet.}
  \label{fig:failure}
  \vspace{10pt}
\end{figure}

\Cref{fig:failure} shows sampled failure cases on ShapeNet.  We categorize those cases into three classes: lack of correspondence, the existence of multiple symmetry planes, and asymmetric objects.  For the first category, e.g., the firearm shown in \Cref{fig:failure}(a), it is hard to accurately find the symmetry plane from the geometry cues because for most pixels, their corresponding points are occluded and invisible in the picture.  For the second category, objects in shapes such as squares and cylinders admit multiple reflection symmetry, and \modelname{} may return the reflection plane that differs from the symmetry plane of the ground truth.  For the third category, some objects in ShapeNet are not symmetric.  Thus, the detected symmetry plane might be different from the ``ground truth symmetry plane'' computed by applying $\Rt$ to the Y-Z plane in the world space.

\subsection{More Visualization}

\begin{figure*}[ht]
  \centering
  \setlength{\lineskip}{0pt}
  \def\mywidth{0.11\linewidth}

\includegraphics[width=\mywidth]{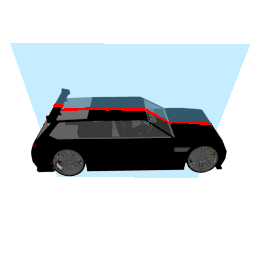}%
\includegraphics[width=\mywidth]{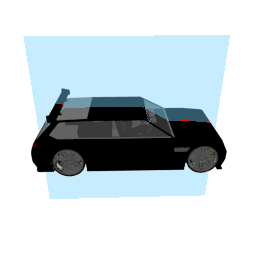}%
\includegraphics[width=\mywidth]{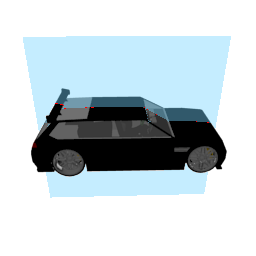}%
\includegraphics[width=\mywidth]{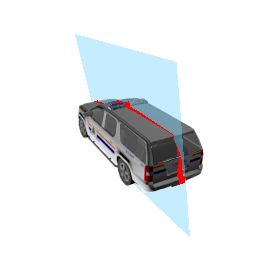}%
\includegraphics[width=\mywidth]{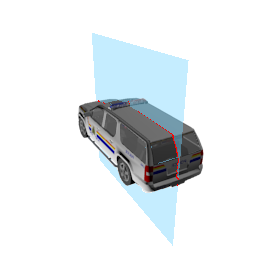}%
\includegraphics[width=\mywidth]{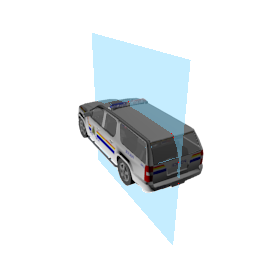}%
\includegraphics[width=\mywidth]{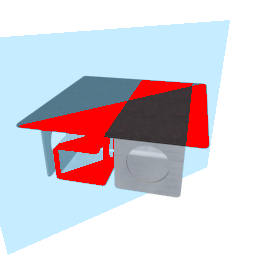}%
\includegraphics[width=\mywidth]{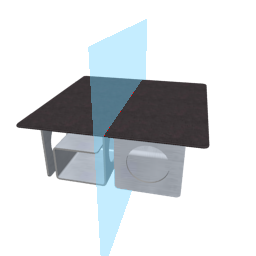}%
\includegraphics[width=\mywidth]{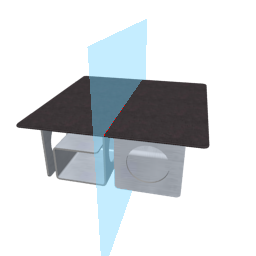}%

\includegraphics[width=\mywidth]{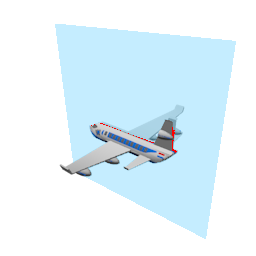}%
\includegraphics[width=\mywidth]{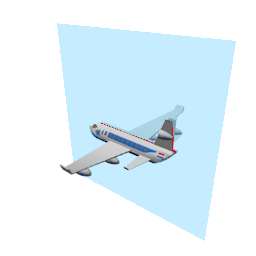}%
\includegraphics[width=\mywidth]{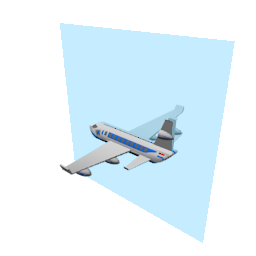}%
\includegraphics[width=\mywidth]{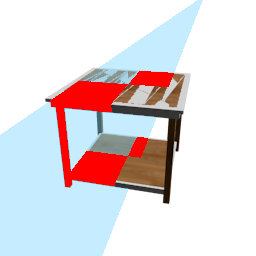}%
\includegraphics[width=\mywidth]{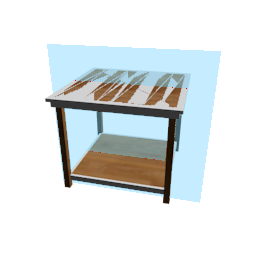}%
\includegraphics[width=\mywidth]{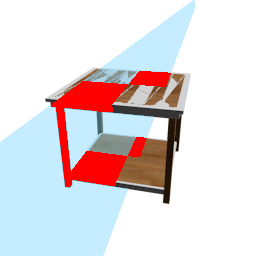}%
\includegraphics[width=\mywidth]{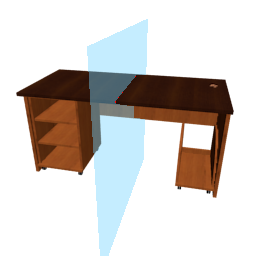}%
\includegraphics[width=\mywidth]{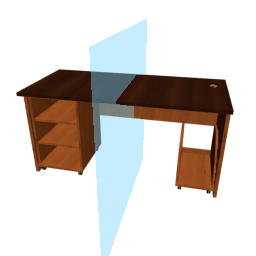}%
\includegraphics[width=\mywidth]{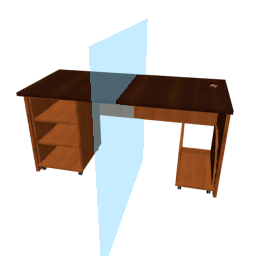}%

\includegraphics[width=\mywidth]{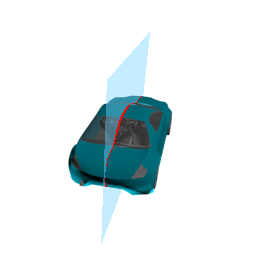}%
\includegraphics[width=\mywidth]{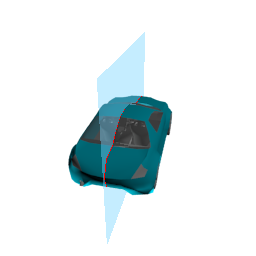}%
\includegraphics[width=\mywidth]{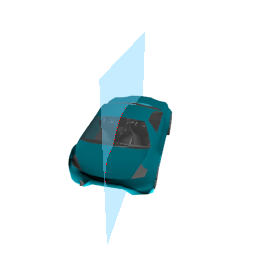}%
\includegraphics[width=\mywidth]{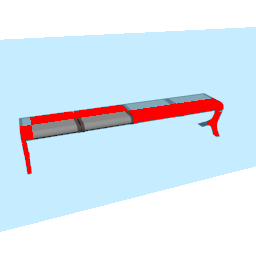}%
\includegraphics[width=\mywidth]{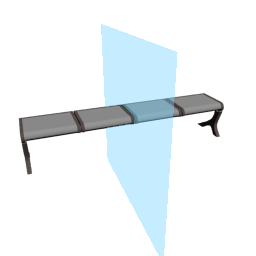}%
\includegraphics[width=\mywidth]{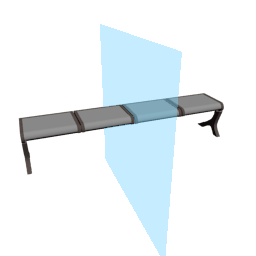}%
\includegraphics[width=\mywidth]{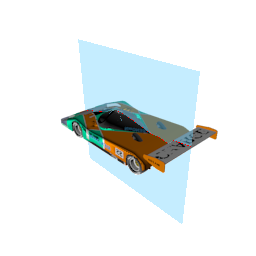}%
\includegraphics[width=\mywidth]{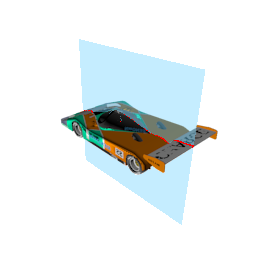}%
\includegraphics[width=\mywidth]{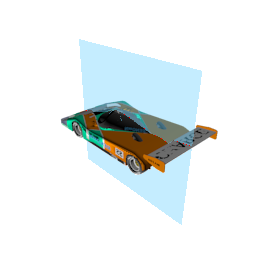}%

\includegraphics[width=\mywidth]{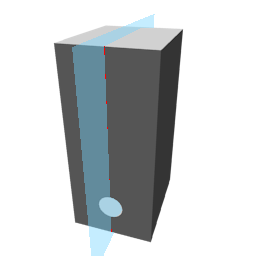}%
\includegraphics[width=\mywidth]{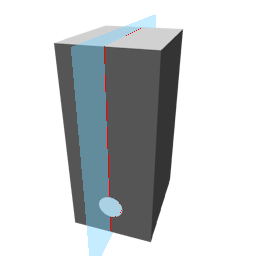}%
\includegraphics[width=\mywidth]{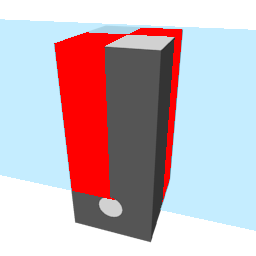}%
\includegraphics[width=\mywidth]{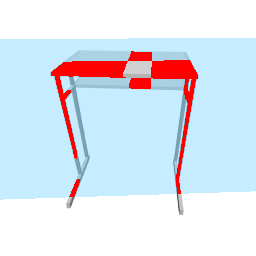}%
\includegraphics[width=\mywidth]{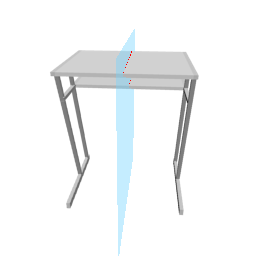}%
\includegraphics[width=\mywidth]{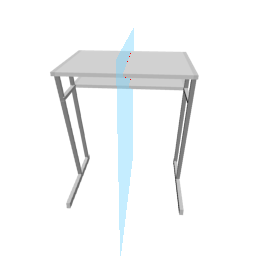}%
\includegraphics[width=\mywidth]{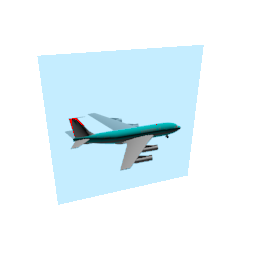}%
\includegraphics[width=\mywidth]{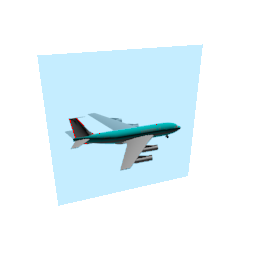}%
\includegraphics[width=\mywidth]{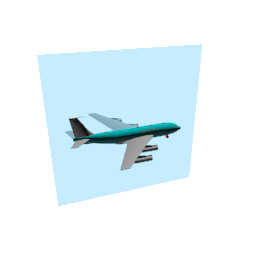}%

\includegraphics[width=\mywidth]{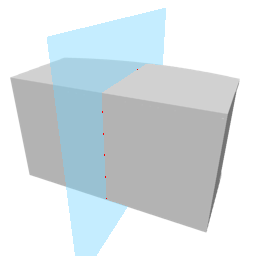}%
\includegraphics[width=\mywidth]{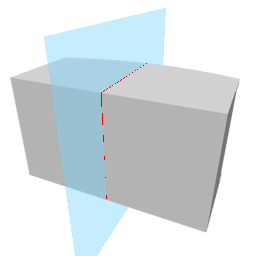}%
\includegraphics[width=\mywidth]{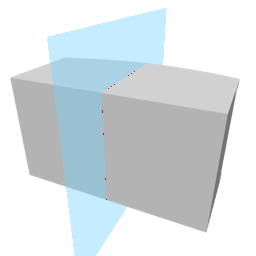}%
\includegraphics[width=\mywidth]{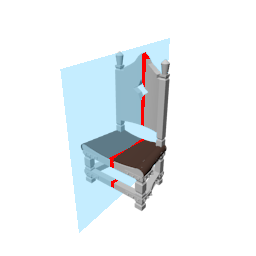}%
\includegraphics[width=\mywidth]{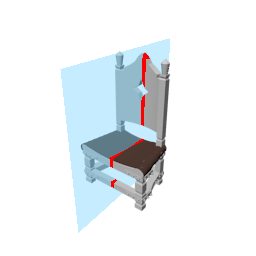}%
\includegraphics[width=\mywidth]{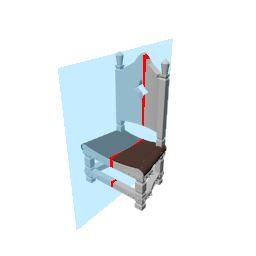}%
\includegraphics[width=\mywidth]{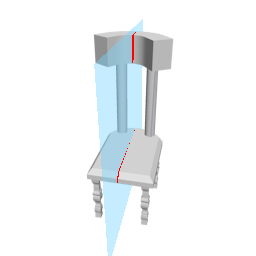}%
\includegraphics[width=\mywidth]{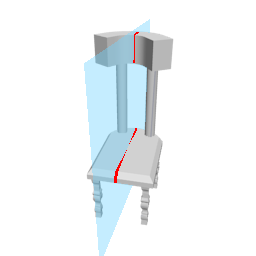}%
\includegraphics[width=\mywidth]{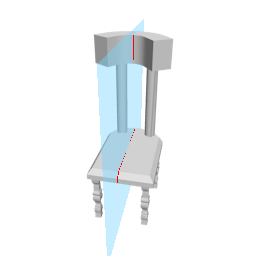}%

\includegraphics[width=\mywidth]{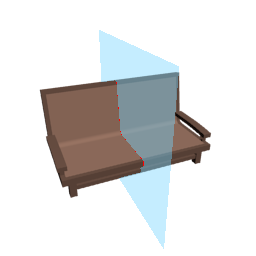}%
\includegraphics[width=\mywidth]{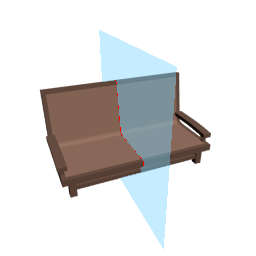}%
\includegraphics[width=\mywidth]{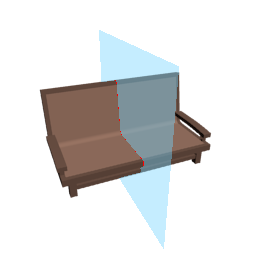}%
\includegraphics[width=\mywidth]{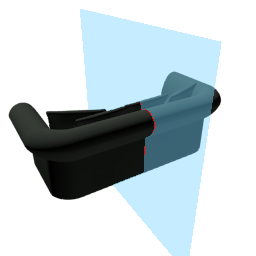}%
\includegraphics[width=\mywidth]{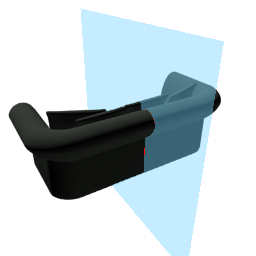}%
\includegraphics[width=\mywidth]{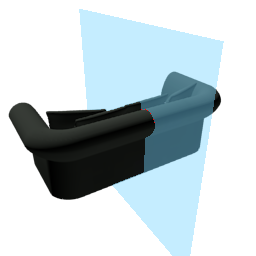}%
\includegraphics[width=\mywidth]{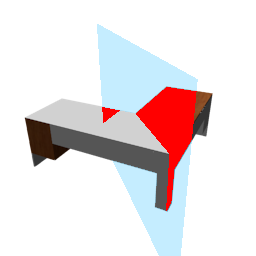}%
\includegraphics[width=\mywidth]{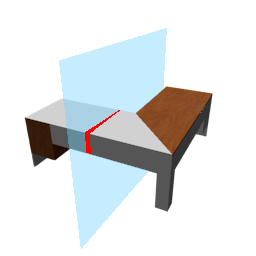}%
\includegraphics[width=\mywidth]{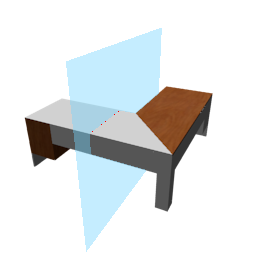}%

\includegraphics[width=\mywidth]{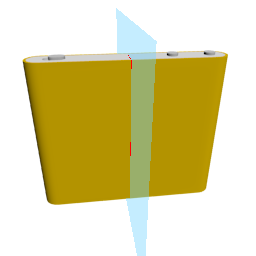}%
\includegraphics[width=\mywidth]{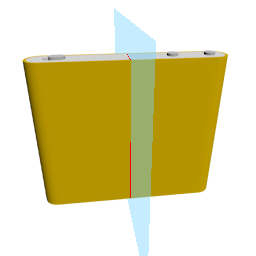}%
\includegraphics[width=\mywidth]{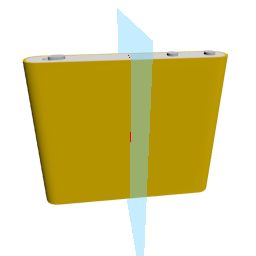}%
\includegraphics[width=\mywidth]{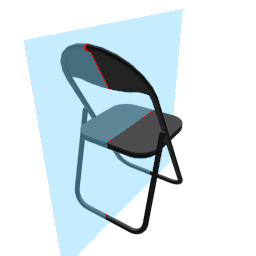}%
\includegraphics[width=\mywidth]{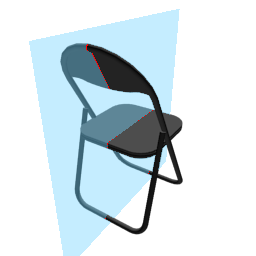}%
\includegraphics[width=\mywidth]{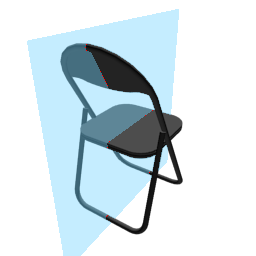}%
\includegraphics[width=\mywidth]{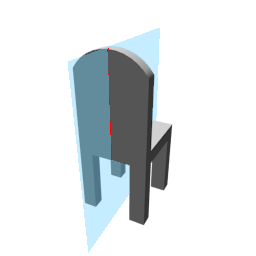}%
\includegraphics[width=\mywidth]{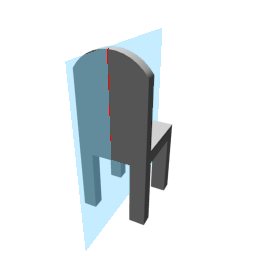}%
\includegraphics[width=\mywidth]{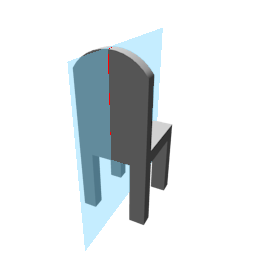}%

\includegraphics[width=\mywidth]{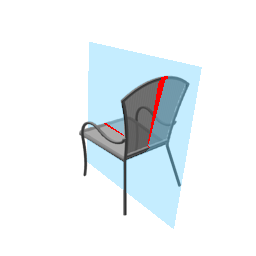}%
\includegraphics[width=\mywidth]{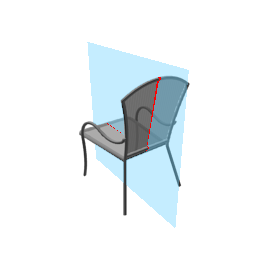}%
\includegraphics[width=\mywidth]{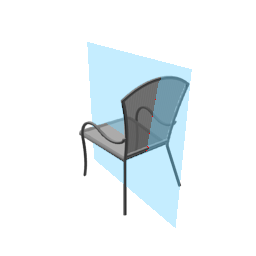}%
\includegraphics[width=\mywidth]{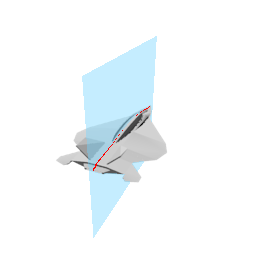}%
\includegraphics[width=\mywidth]{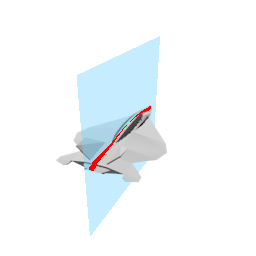}%
\includegraphics[width=\mywidth]{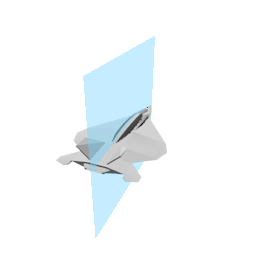}%
\includegraphics[width=\mywidth]{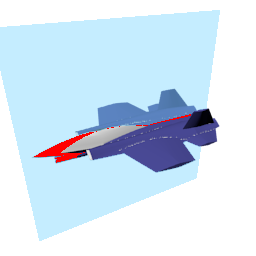}%
\includegraphics[width=\mywidth]{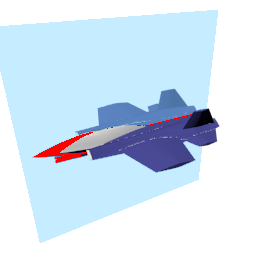}%
\includegraphics[width=\mywidth]{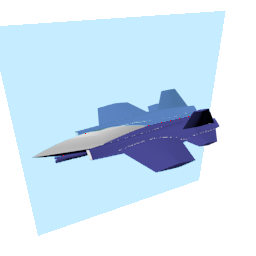}%

\includegraphics[width=\mywidth]{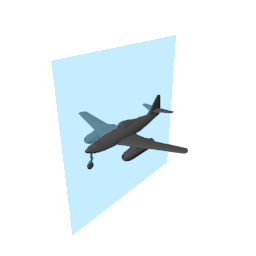}%
\includegraphics[width=\mywidth]{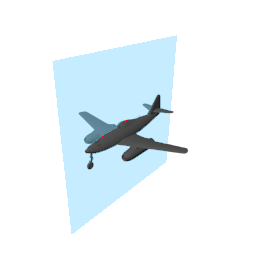}%
\includegraphics[width=\mywidth]{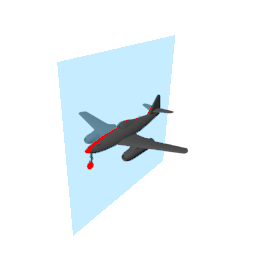}%
\includegraphics[width=\mywidth]{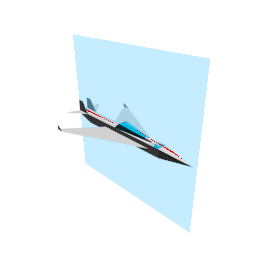}%
\includegraphics[width=\mywidth]{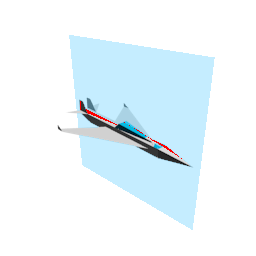}%
\includegraphics[width=\mywidth]{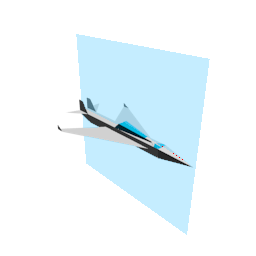}%
\includegraphics[width=\mywidth]{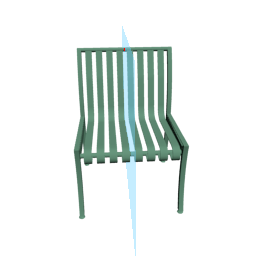}%
\includegraphics[width=\mywidth]{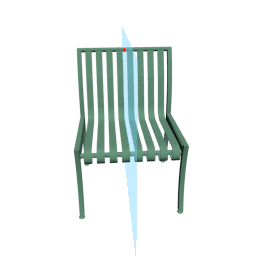}%
\includegraphics[width=\mywidth]{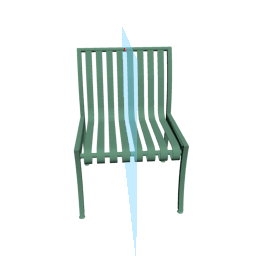}%

\includegraphics[width=\mywidth]{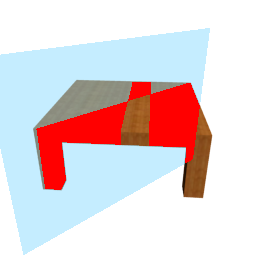}%
\includegraphics[width=\mywidth]{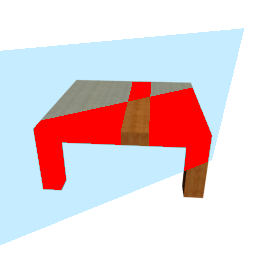}%
\includegraphics[width=\mywidth]{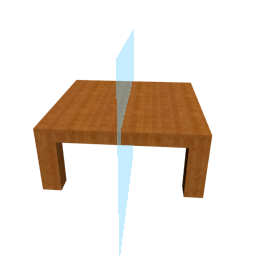}%
\includegraphics[width=\mywidth]{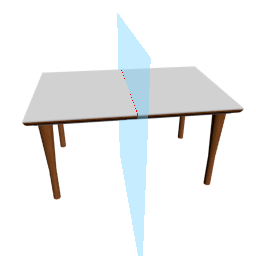}%
\includegraphics[width=\mywidth]{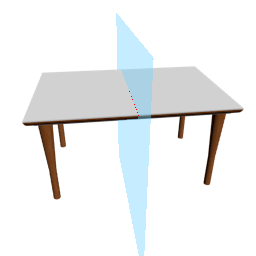}%
\includegraphics[width=\mywidth]{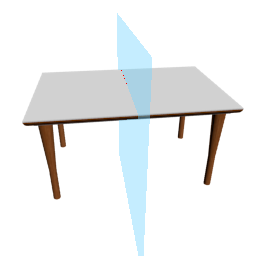}%
\includegraphics[width=\mywidth]{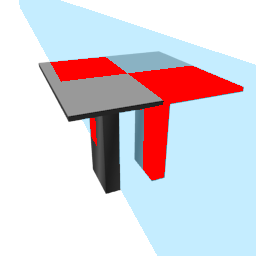}%
\includegraphics[width=\mywidth]{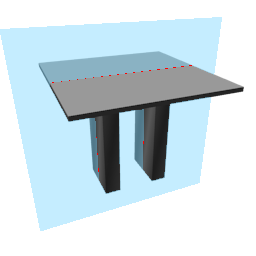}%
\includegraphics[width=\mywidth]{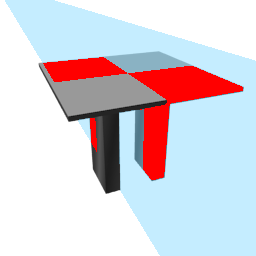}%

\includegraphics[width=\mywidth]{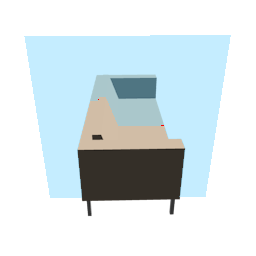}%
\includegraphics[width=\mywidth]{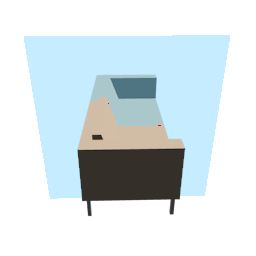}%
\includegraphics[width=\mywidth]{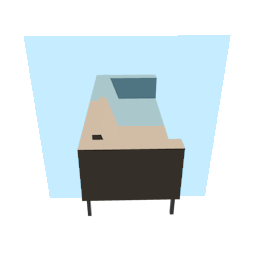}%
\includegraphics[width=\mywidth]{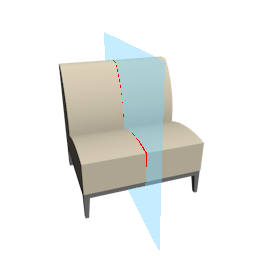}%
\includegraphics[width=\mywidth]{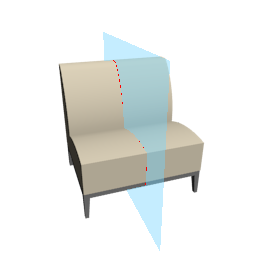}%
\includegraphics[width=\mywidth]{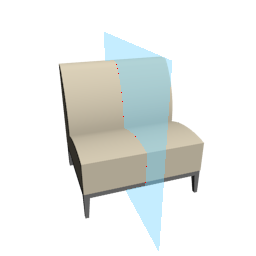}%
\includegraphics[width=\mywidth]{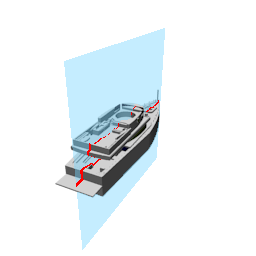}%
\includegraphics[width=\mywidth]{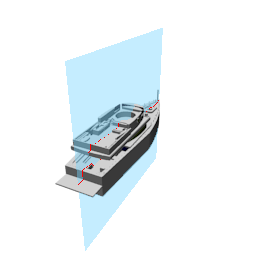}%
\includegraphics[width=\mywidth]{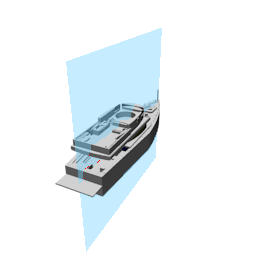}%

\small
\begin{minipage}[t]{\mywidth}\centering Front2Back\end{minipage}%
\begin{minipage}[t]{\mywidth}\centering ResNet\end{minipage}%
\begin{minipage}[t]{\mywidth}\centering \modelname{}\end{minipage}%
\begin{minipage}[t]{\mywidth}\centering Front2Back\end{minipage}%
\begin{minipage}[t]{\mywidth}\centering ResNet\end{minipage}%
\begin{minipage}[t]{\mywidth}\centering \modelname{}\end{minipage}%
\begin{minipage}[t]{\mywidth}\centering Front2Back\end{minipage}%
\begin{minipage}[t]{\mywidth}\centering ResNet\end{minipage}%
\begin{minipage}[t]{\mywidth}\centering \modelname{}\end{minipage}%
  \vspace{5pt}

  \caption{Detected symmetry planes of \modelname{} on \emph{random sampled images} from ShapeNet.  Errors of symmetry planes, i.e., pixels between the predicted plane and the ground truth plane, are \textcolor{red}{highlighted}.}
  \label{fig:supp:shapenet:planes}
\end{figure*}

\begin{figure*}[ht]
  \centering
  \setlength{\lineskip}{8pt}
  \def\mywidth{0.10\linewidth}

\hfill\includegraphics[width=\mywidth]{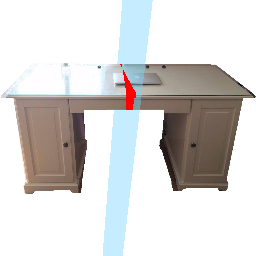}%
\hfill\includegraphics[width=\mywidth]{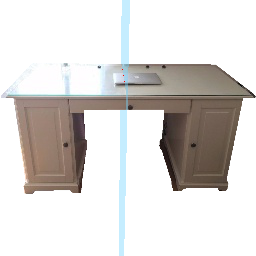}%
\hfill\includegraphics[width=\mywidth]{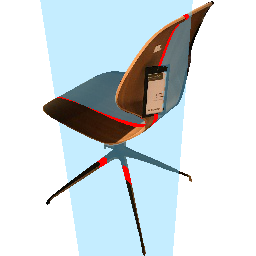}%
\hfill\includegraphics[width=\mywidth]{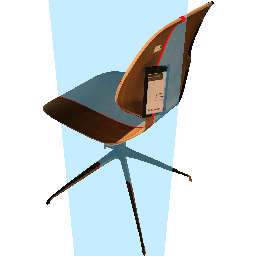}%
\hfill\includegraphics[width=\mywidth]{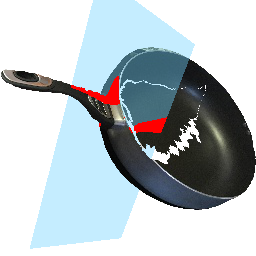}%
\hfill\includegraphics[width=\mywidth]{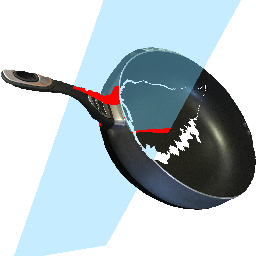}%
\hfill\includegraphics[width=\mywidth]{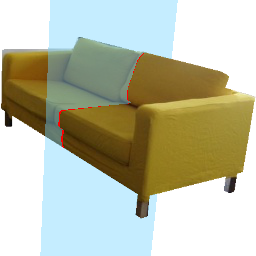}%
\hfill\includegraphics[width=\mywidth]{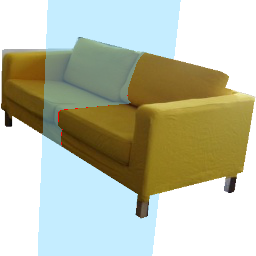}%
\hfill

\hfill\includegraphics[width=\mywidth]{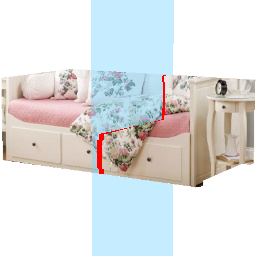}%
\hfill\includegraphics[width=\mywidth]{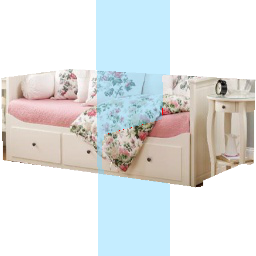}%
\hfill\includegraphics[width=\mywidth]{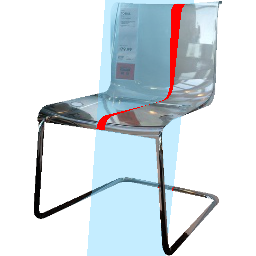}%
\hfill\includegraphics[width=\mywidth]{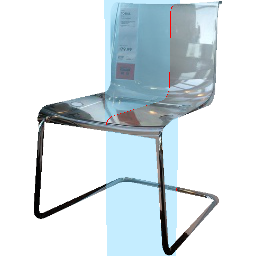}%
\hfill\includegraphics[width=\mywidth]{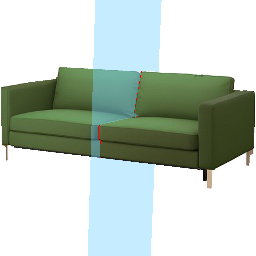}%
\hfill\includegraphics[width=\mywidth]{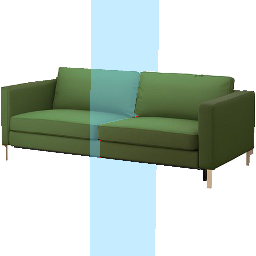}%
\hfill\includegraphics[width=\mywidth]{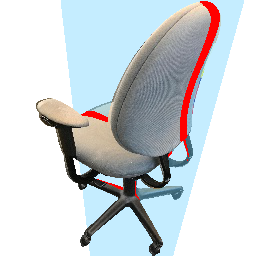}%
\hfill\includegraphics[width=\mywidth]{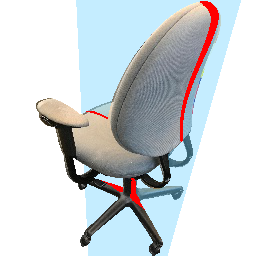}%
\hfill

\hfill\includegraphics[width=\mywidth]{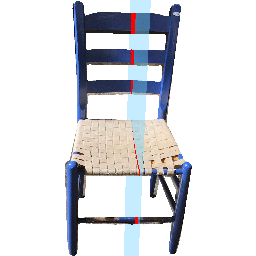}%
\hfill\includegraphics[width=\mywidth]{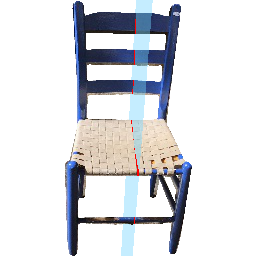}%
\hfill\includegraphics[width=\mywidth]{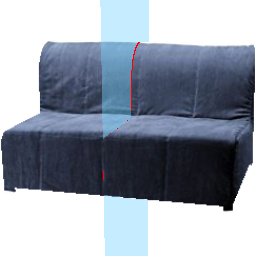}%
\hfill\includegraphics[width=\mywidth]{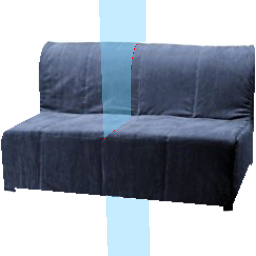}%
\hfill\includegraphics[width=\mywidth]{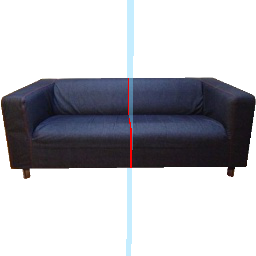}%
\hfill\includegraphics[width=\mywidth]{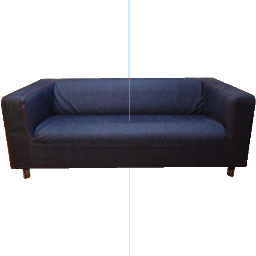}%
\hfill\includegraphics[width=\mywidth]{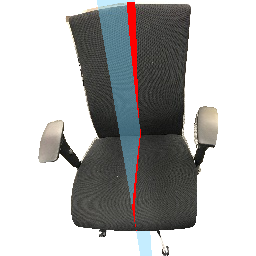}%
\hfill\includegraphics[width=\mywidth]{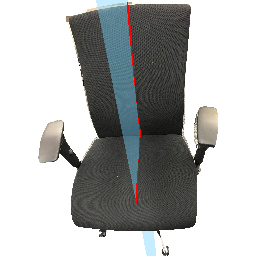}%
\hfill

\hfill\includegraphics[width=\mywidth]{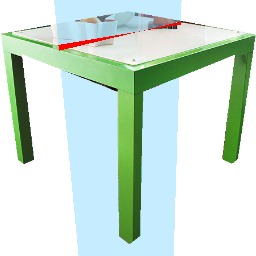}%
\hfill\includegraphics[width=\mywidth]{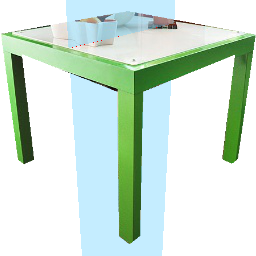}%
\hfill\includegraphics[width=\mywidth]{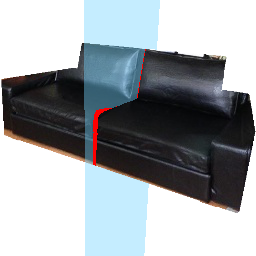}%
\hfill\includegraphics[width=\mywidth]{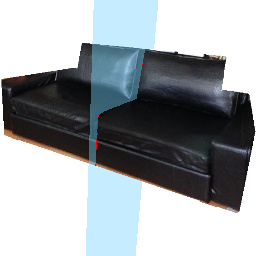}%
\hfill\includegraphics[width=\mywidth]{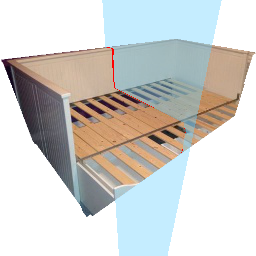}%
\hfill\includegraphics[width=\mywidth]{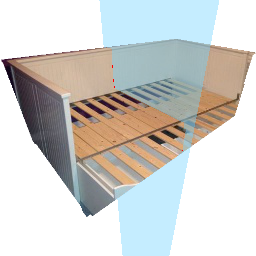}%
\hfill\includegraphics[width=\mywidth]{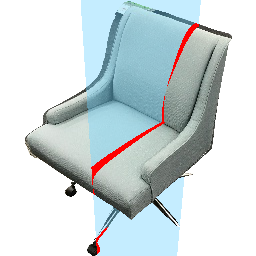}%
\hfill\includegraphics[width=\mywidth]{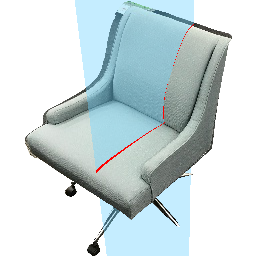}%
\hfill

\hfill\includegraphics[width=\mywidth]{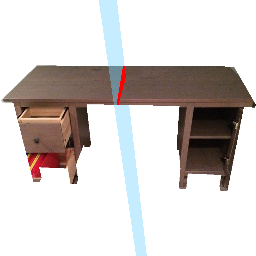}%
\hfill\includegraphics[width=\mywidth]{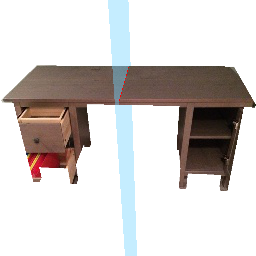}%
\hfill\includegraphics[width=\mywidth]{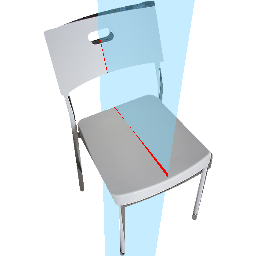}%
\hfill\includegraphics[width=\mywidth]{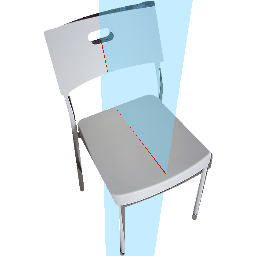}%
\hfill\includegraphics[width=\mywidth]{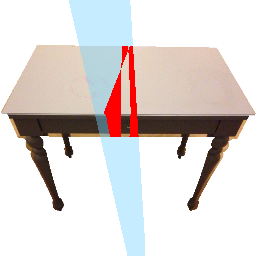}%
\hfill\includegraphics[width=\mywidth]{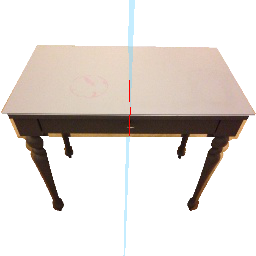}%
\hfill\includegraphics[width=\mywidth]{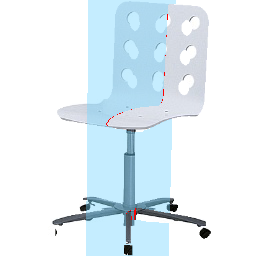}%
\hfill\includegraphics[width=\mywidth]{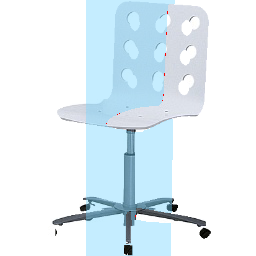}%
\hfill

\hfill\includegraphics[width=\mywidth]{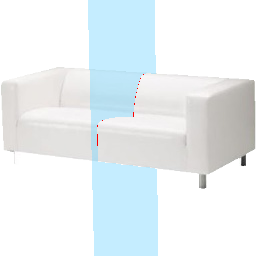}%
\hfill\includegraphics[width=\mywidth]{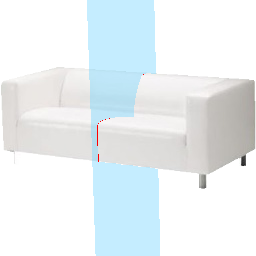}%
\hfill\includegraphics[width=\mywidth]{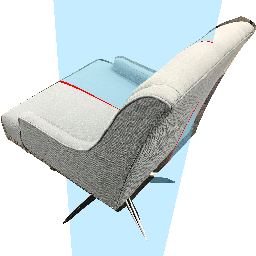}%
\hfill\includegraphics[width=\mywidth]{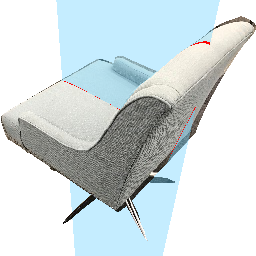}%
\hfill\includegraphics[width=\mywidth]{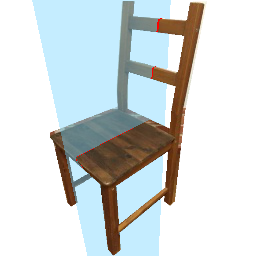}%
\hfill\includegraphics[width=\mywidth]{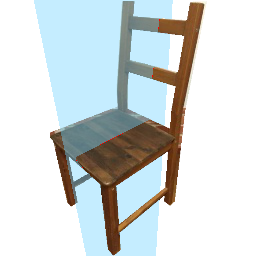}%
\hfill\includegraphics[width=\mywidth]{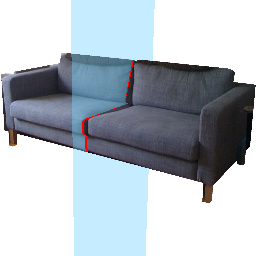}%
\hfill\includegraphics[width=\mywidth]{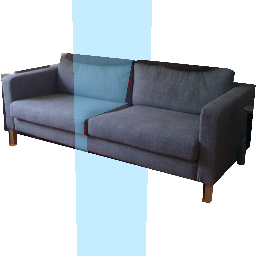}%
\hfill

\hfill\includegraphics[width=\mywidth]{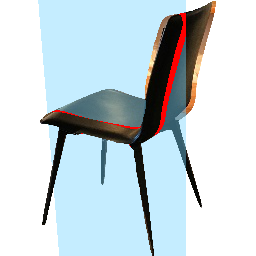}%
\hfill\includegraphics[width=\mywidth]{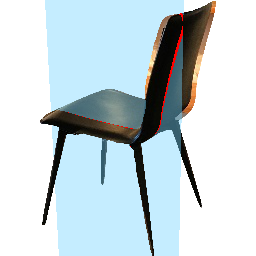}%
\hfill\includegraphics[width=\mywidth]{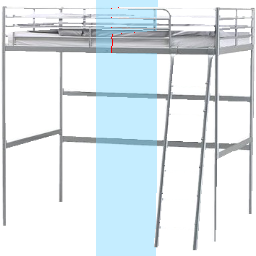}%
\hfill\includegraphics[width=\mywidth]{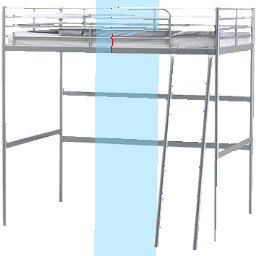}%
\hfill\includegraphics[width=\mywidth]{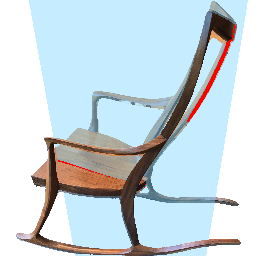}%
\hfill\includegraphics[width=\mywidth]{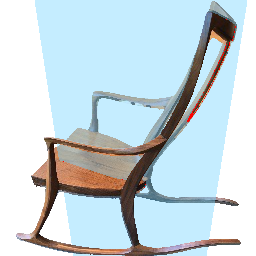}%
\hfill\includegraphics[width=\mywidth]{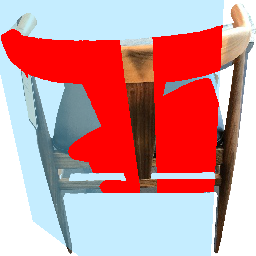}%
\hfill\includegraphics[width=\mywidth]{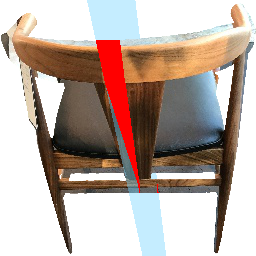}%
\hfill

\hfill\includegraphics[width=\mywidth]{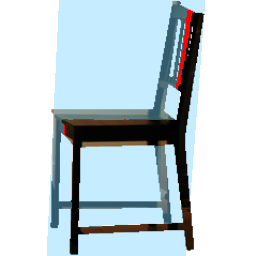}%
\hfill\includegraphics[width=\mywidth]{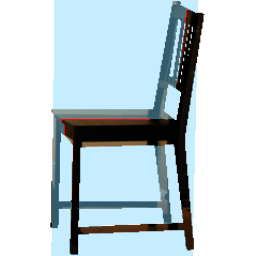}%
\hfill\includegraphics[width=\mywidth]{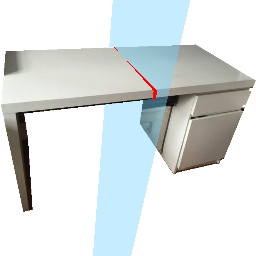}%
\hfill\includegraphics[width=\mywidth]{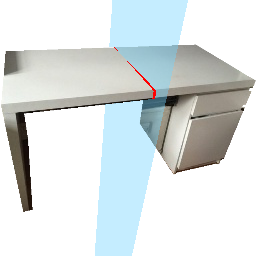}%
\hfill\includegraphics[width=\mywidth]{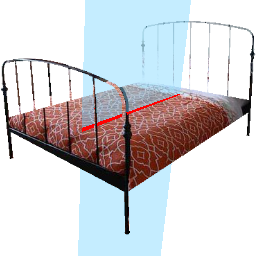}%
\hfill\includegraphics[width=\mywidth]{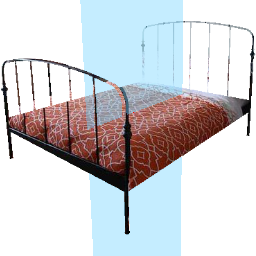}%
\hfill\includegraphics[width=\mywidth]{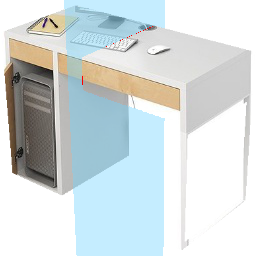}%
\hfill\includegraphics[width=\mywidth]{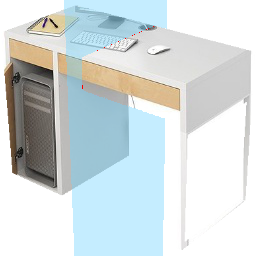}%
\hfill

\hfill\includegraphics[width=\mywidth]{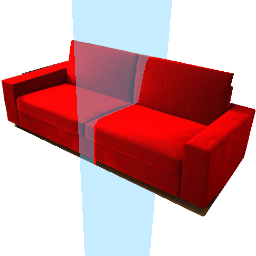}%
\hfill\includegraphics[width=\mywidth]{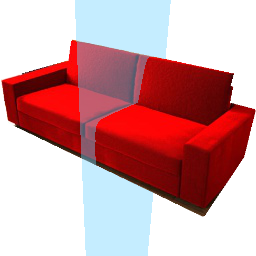}%
\hfill\includegraphics[width=\mywidth]{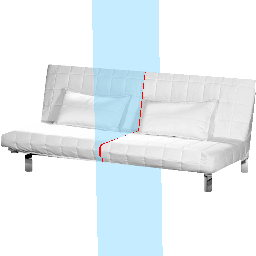}%
\hfill\includegraphics[width=\mywidth]{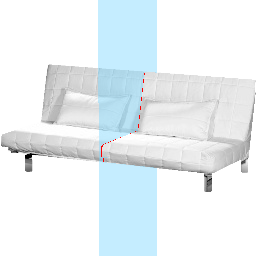}%
\hfill\includegraphics[width=\mywidth]{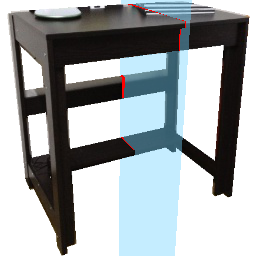}%
\hfill\includegraphics[width=\mywidth]{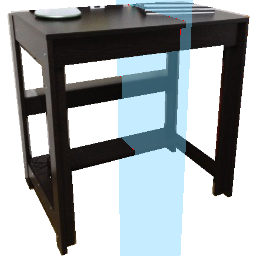}%
\hfill\includegraphics[width=\mywidth]{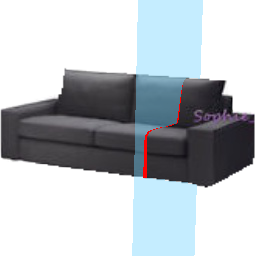}%
\hfill\includegraphics[width=\mywidth]{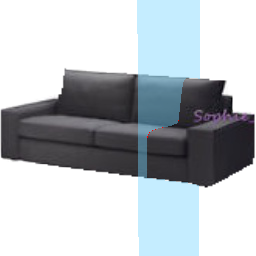}%
\hfill

\hfill\includegraphics[width=\mywidth]{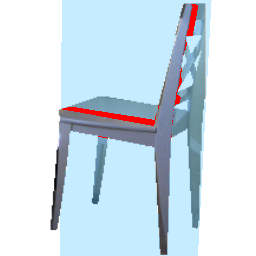}%
\hfill\includegraphics[width=\mywidth]{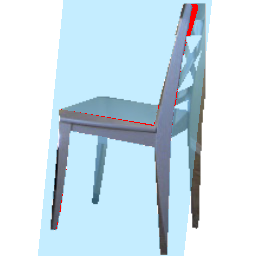}%
\hfill\includegraphics[width=\mywidth]{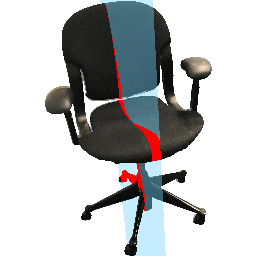}%
\hfill\includegraphics[width=\mywidth]{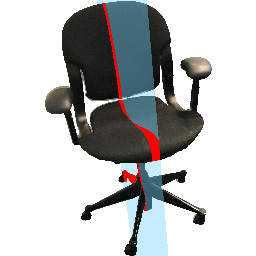}%
\hfill\includegraphics[width=\mywidth]{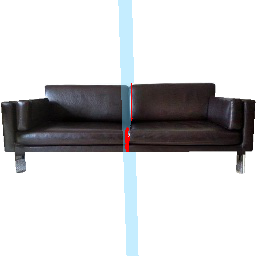}%
\hfill\includegraphics[width=\mywidth]{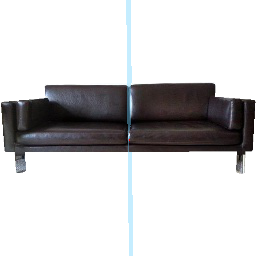}%
\hfill\includegraphics[width=\mywidth]{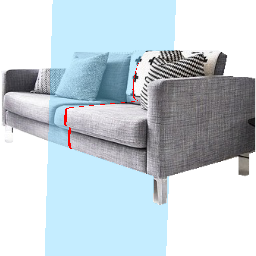}%
\hfill\includegraphics[width=\mywidth]{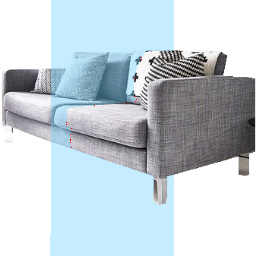}%
\hfill

  \vspace{5pt}
\small
\hfill\begin{minipage}[t]{\mywidth}\centering ResNet\end{minipage}%
\hfill\begin{minipage}[t]{\mywidth}\centering \modelname{}\end{minipage}%
\hfill\begin{minipage}[t]{\mywidth}\centering ResNet\end{minipage}%
\hfill\begin{minipage}[t]{\mywidth}\centering \modelname{}\end{minipage}%
\hfill\begin{minipage}[t]{\mywidth}\centering ResNet\end{minipage}%
\hfill\begin{minipage}[t]{\mywidth}\centering \modelname{}\end{minipage}%
\hfill\begin{minipage}[t]{\mywidth}\centering ResNet\end{minipage}%
\hfill\begin{minipage}[t]{\mywidth}\centering \modelname{}\end{minipage}%
  \vspace{5pt}

  \caption{Detected symmetry planes of \modelname{} on \emph{random sampled images} from Pix3D.  Errors of symmetry planes, i.e., pixels between the predicted plane and the ground truth plane, are \textcolor{red}{highlighted}. Artifacts in visualization are due to the misalignment of between the models and images in Pix3D.}
  \label{fig:supp:pix3d:planes}
\end{figure*}

\Cref{fig:supp:shapenet:planes} and \Cref{fig:supp:pix3d:planes} show the visual quality of the detected symmetry planes of \modelname{}  on \textbf{random sampled} testing images from ShapeNet and Pix3D, respectively.  For the results of Pix3D, we note that artifacts in visualization such as acentric symmetry planes are due to misalignment of CAD models and real-world images.  We find that for most images, \modelname{} is able to determine the normal of symmetry plane accurately by utilization the geometric constraints from symmetry.  The errors (red pixels) are sub-pixel in most cases.

\end{document}